\documentclass[10pt,a4paper,journal,final]{IEEEtran}
\usepackage[latin9]{inputenc}
\usepackage{url}
\usepackage{amsmath}
\usepackage{amssymb}
\usepackage{graphicx}

\makeatletter

\providecommand{\tabularnewline}{\\}

\usepackage{url}
\usepackage{amsfonts}
\usepackage{rotating}
\providecommand{\U}[1]{\protect\rule{.1in}{.1in}}
\providecommand{\tabularnewline}{\\}

\DeclareTextSymbolDefault{\textquotedbl}{T1}
\providecommand{\tabularnewline}{\\}

\providecommand{\tabularnewline}{\\}

\providecommand{\tabularnewline}{\\}

\providecommand{\tabularnewline}{\\}

\makeatother

\begin{document}
\title{How Far Two UAVs Should Be subject to Communication Uncertainties}
\author{Quan~Quan, Rao Fu, and Kai-Yuan Cai\thanks{Q. Quan, R. Fu and K-Y. Cai are with the School of Automation Science
and Electrical Engineering, Beihang University, Beijing 100191, China
(e-mail: qq\_buaa@buaa.edu.cn; buaafurao@buaa.edu.cn; kycai@buaa.edu.cn).}}
\maketitle
\begin{abstract}
Unmanned aerial vehicles are now becoming increasingly accessible
to amateur and commercial users alike. A safety air traffic management
system is needed to help ensure that every newest entrant into the
sky does not collide with others. Much research has been done to design
various methods to perform collision avoidance with obstacles. However,
how to decide the safety radius subject to communication uncertainties
is still suspended. Based on assumptions on communication uncertainties
and supposed control performance, a separation principle of the safety
radius design and controller design is proposed. With it, the safety
radius corresponding to the safety area in the design phase (without
uncertainties) and flight phase (subject to uncertainties) are studied.
Furthermore, the results are extended to multiple obstacles. Simulations
and experiments are carried out to show the effectiveness of the proposed
methods.
\end{abstract}


\section{Introduction}

With the on-going miniaturization of motors, sensors and processors,
the number of Unmanned aerial vehicles(UAVs) continues to explode
and they increasingly play an integral role in many practical applications
in places where the working environment is dangerous or human capacity
is limited \cite{Balakrishnan(2018)}. Hence, in increasingly busy
airspace, the conflicts among UAVs will occur frequently and become
a serious problem. UAS Traffic Management (UTM) by NASA in the USA
\cite{NASA} and U-SPACE in the European Union \cite{SESAR} in progress
are aiming to manage UAVs for tactical self-separation and collision
avoidance \cite{Depoorter(2019)}.

Traditionally, the main role of air traffic management (ATM) is to
keep a prescribed separation between all aircraft by using centralized
control. However, it is infeasible for increasing UAVs because the
traditional control method lacks scalability. In order to address
such a problem, free flight is a developing air traffic control method
that uses decentralized control \cite{FreeFlight(1995)},\cite{Jana(1997)}.
By Automatic Dependent Surveillance-Broadcast (ADS-B) \cite{ICAO(2012)},
Vehicle to Vehicle (V2V) communication \cite{Chakrabarty(2019)}\ or
5G mobile network \cite{Hayat(2016)}, UAVs can receive the information
of their neighboring obstacles to avoid collision. Especially in low-altitude
airspace, utilizing the existing mobile networks will eliminate the
need to deploy new infrastructure and, therefore, help to ensure connected
UAVs are economically feasible \cite{GSMA(2018)}. Communication is
often considered for UAVs, especially in the formation control, where
the communication connection establishes the network topological structure
\cite{Dong(2015)},\cite{Seo(2018)},\cite{Dutta(2018)}. However,
it is not enough because communication protocols are subject to message
delivery delay and packet loss. A V2V communication test on UAVs has
been done by NASA for UTM showing that the probability of packet loss
will be increased gradually to $1$ after two UAVs keep about $1.3$km
away \cite{Glaab(2018)}. A UAV typically relies on localization sensors,
monitoring systems, and/or wireless communication networks. These
sensing mechanisms will provide inaccurate position information as
a result of process delays, interferences, noise, and quantization,
jeopardizing the safety of the vehicle and its environment \cite{Rodriguez-Seda(2016)}.
If these uncertainties are not adequately taken into consideration,
then UAVs may become vulnerable to collisions. This motivates us to
study UAV collision avoidance problem for moving obstacles \cite{Wang(2007)},\cite{Haugen(2016)}
subject to sensing uncertainties, especially \emph{communication uncertainties}.

\begin{figure}[h]
\begin{centering}
\includegraphics[scale=0.4]{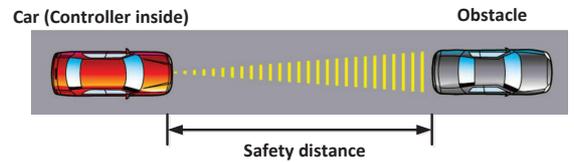}
\par\end{centering}
\caption{Safe distance of cars in highway.}
\label{carexample}
\end{figure}

Much research has been done to design various methods, including path
planning, conflict resolution, model predictive control, potential
field, geometric guidance, motion planning of teams, for UAVs to perform
collision avoidance with obstacles \cite{Huang(2019)},\cite{Mcfadyen(2016)}.
A simulation study of four typical collision avoidance methods can
be found in \cite{Mueller(2016)}. Most of existing collision avoidance
methods suppose that they can acquire exact sensing information. However,
in recent years, more and more attention has been paid to collision
avoidance subject to sensing uncertainties. Two ways are followed
to extend the existing collision avoidance methods to handle sensing
uncertainties for moving obstacles.
\begin{itemize}
\item The major way is to predict the future obstacles' trajectory set with
uncertainty models \cite{Bai(2011)}, \cite{Yang(2016)}, \cite{Zhou(2019)},\cite{Arul(2019)},\cite{Angeris(2019)},\cite{Zhu(2019)}.
Then, control, decision or planning is obtained by optimization over
a future time horizon to avoid obstacles' trajectory sets or reachable
sets in the sense of probability. This way is often for the control
methods without closed forms, like the optimization methods mentioned
above. However, as pointed in \cite{Alonso-Mora(2018)}, under some
circumstances the optimization problem may be infeasible due to the
sensing uncertainties. Moreover, heavy computational burden always
introduces difficulties in the design process. A discussion will be
given to show this in Section IV.C. As far as we know, fewer communication
uncertainties are considered in this way.
\item The other way is to take the sensing uncertainties into the analysis
of the closed-loop of existing methods. Then, adjust controller parameters
to reject uncertainties \cite{Rodriguez-Seda(2016)}. This way is
often for the control methods with closed forms, like the potential
field method. However, the analysis will be more difficult subject
to the uncertainties like communication uncertainties. What is worse,
the communication uncertainties may make the closed-loop instability
if the delay or packet loss is not compensated for elaborately \cite{Yamchi(2017)}. 
\end{itemize}
As pointed out in \cite{Viragh(2016)}, communication (delivery delay
and packet loss) is a very significant and necessary factor that should
be considered in UAV swarm. The survey paper \cite{Huang(2019)} also
takes communication uncertainties in collision avoidance as a challenge.
To deal with sensing uncertainties including communication uncertainties,
\emph{a complementary way of collision avoidance is proposed in this
paper with a principle that separates the safety radius design for
uncertainties and controller design for collision avoidance}. Here,
the safety radius design will take all sensing uncertainties into
consideration, while the controller design does not need to take uncertainties
into consideration. So, the two ways mentioned above can also benifit
from the safety radius design when facing communication uncertainties.
Intuitively, the safety radius will be increased as the uncertainties
are increased. This is inspired by traffic rules in both the aviation
area \cite{Hoekstra(2002)} and the ground transportation area \cite{Iranmanesh(2016)}.
For example, two airplanes should maintain standard en route separation
between aircraft (5 nautical miles (9.3 km) horizontal and 1,000 feet
(300 m) vertical) \cite{Hoekstra(2002)}. Also, as shown in Figure
\ref{carexample}, it is well known that two cars on highway should
keep a certain safe distance. However, the experience to determine
the safety radius for manned and airline airplanes in high-altitude
airspace or safety distance for cars is difficult or not at all to
apply to decentralized-control UAVs in low-altitude space because
of the differences in pilot manner, communication manner, flight manner
and risk requirement. On the other hand, obviously, the safety radius
cannot be just the physical radius of a UAV because many uncertainties,
like estimation error, communication delay and packet loss, should
be considered. As pointed out in \cite{Huang(2019)}, the safety radius
design is a challenge in the presence of sensing uncertainties, especially
communication uncertainties.

To this end, this paper focuses on studying the safety radius of decentralized-control
Vertical TakeOff and Landing (VTOL) UAVs, while the deterministic
collision avoidance controller could be any type as long as it satisfies
certain conditions. VTOL ability, which enables easier grounding or
holding by hovering, is an important ability that might be mandated
by authorities in high traffic areas such as lower altitude in the
urban airspace \cite{IoD(2016)}. This is because VTOL drones are
highly versatile and can perform tasks in an environment with very
little available airspace. The study on the safety radius is divided
into two phases, namely the `offline' design phase and `online' flight
phase.
\begin{itemize}
\item In the design phase, when a UAV is on the ground, a question will
arise that \emph{how far} the UAV and an obstacle should be kept without
uncertainties in order to avoid a collision in the presence of uncertainties?
\item On the other hand, when a UAV in the sky (in the presence of uncertainties),
a question will arise that \emph{how far} should the UAV and an obstacle
be kept in the sense of estimated distance (involve uncertainties
but only can be accessed) in order to avoid a collision? 
\end{itemize}
To reply to these two questions, first, a VTOL UAV control model and
an obstacle model are proposed. The filtered position is defined to
replace position by considering velocity. Based on these models above,
assumptions on practical uncertainties, like estimation error, communication
delay and packet loss, are considered in the broadcast information
received by the UAV. Assumptions on control performance are assumed
for the design phase and the flight phase, respectively. Since only
is the distance error used, the controller can be distributed. Based
on these assumptions, a principle of separation of control and safety
radius is proposed in the design phase (\textit{Theorem 1}). Based
on this principle, the safety radius in the design phase is derived.
Then, the safety radius in the flight phase is further derived in
the sense of estimated distance. These conclusions are further extended
to multiple obstacles. Simulations and experiments including delay
and package-loss uncertainties are carried out to show the effectiveness
of the proposed method. 

As shown in Figure \ref{carexample}, let us recall the safe distance
for cars in highway (1D space), where drivers in cars are controllers.
When set the safety distance for all of us, we in fact make an assumption
on the control performance for all drivers (controllers) by taking
the drivers' response delay and road condition's uncertainties. This
idea leaves uncertainties in the safe distance. Motivated by this,
we do the similar work here for UAVs in 3D space. In the past decades,
many collision-avoidance controllers have been proposed \cite{VDBerg(2011)},\cite{Fox(1997)},\cite{Richards(2002)}
similar to different drivers. We only need to focus on their performance
rather than what they are. Compared with the safe distance in the
traditional manned traffic field, there are two extras taken into
consideration: i) cooperative obstacles, which will make collision
avoidance at the same time; ii) the packet loss, namely a dynamic
uncertainty. These make the obtained conclusion not easy. The major
contribution of this paper is to separate the controller design and
the safety radius so that the existing methods are available again
(not need to be changed) in the presence of communication uncertainties.
Concretely, the contributions of this paper are: i) a principle of
separation of control and safety radius proposed so that the same
controller with different safety radiuses can deal with different
communication uncertainties; ii) safety radius derived in both the
design phase and the flight phase to reduce the conservatism as much
as possible; iii) filtered VTOL UAV control model in the form of a
single integrator which takes the maneuverability into consideration.

The remainder of this paper is organized as follows. In Section II,
the problem is formulated, which based on a proposed VTOL UAV control
model, a proposed obstacle model, and assumptions. The solutions to
the problem and their extensions are proposed in Section III. The
effectiveness of the proposed safety radius design is demonstrated
by simulation and flight experiments in Section IV. The conclusions
are given in Section V. Some details of the mathematical proof process
are given in Section VI as an appendix.

\section{Problem Formulation}

In this section, a VTOL UAV model and obstacle model are introduced
first. Then, assumptions on uncertainties and control performance
are proposed. Based on models and assumptions, problems for the design
phase objective and flight phase objective are formulated.

\subsection{VTOL UAV Control\ Model}

In a local airspace, there exists a VTOL UAV defined as 
\begin{equation}\setlength{\abovedisplayskip}{4pt}
\setlength{\belowdisplayskip}{4pt}
\mathcal{U}=\left\{ \mathbf{x}\in\mathbb{R}^{3}\left\vert \left\Vert \mathbf{x}-\mathbf{p}\right\Vert <r_{\text{m}}\right.\right\} \label{Physicalaera}
\end{equation}
where $r_{\text{m}}>0$ is the \emph{physical} \emph{radius} of the
UAV related to its physical size, and $\mathbf{p}\in\mathbb{R}^{3}$
is the center of mass of the UAV. Many organizations or companies
have designed some open-source semi-autonomous autopilots or offered
semi-autonomous autopilots with Software Development Kits. The semi-autonomous
autopilots can be used for velocity control of VTOL UAVs. With such
an autopilot, the velocity of a VTOL UAV can track a given velocity
command in a reasonable time. It can not only avoid the trouble of
modifying the low-level source code of autopilots but also utilize
commercial autopilots to complete various tasks. Based on this, the
control model of the UAV\ satisfies 
\begin{align}\setlength{\abovedisplayskip}{4pt}
\setlength{\belowdisplayskip}{4pt}
\mathbf{\dot{p}} & =\mathbf{v}\nonumber \\
\mathbf{\dot{v}} & =-l\left(\mathbf{v}-\mathbf{v}_{\text{c}}\right)\label{positionmodel_ab_con_i}
\end{align}
where $l>0,$ $\mathbf{v}\in{{\mathbb{R}}^{3}}$ is the velocity,
and $\mathbf{v}_{\text{c}}\in{{\mathbb{R}}^{3}}$ is the velocity
command. From the model (\ref{positionmodel_ab_con_i}), $\underset{t\rightarrow\infty}{\lim}\left\Vert \mathbf{v}\left(t\right)-\mathbf{v}_{\text{c}}\right\Vert =0$
if $\mathbf{v}_{\text{c}}$ is constant. The constant $l,$ called
\emph{maneuver constant} here, depends on the VTOL UAV and the semi-autonomous
autopilot used, which can be obtained through flight experiments.
It stands for the maneuverability of the VTOL UAV. If it is big, then
$\mathbf{v}$ can converge to $\mathbf{v}_{\text{c}}$ rapidly, vice
versa. Here, the velocity command $\mathbf{v}_{\text{c}}$ (required
to design) for the VTOL UAV is subject to 
\begin{equation}\setlength{\abovedisplayskip}{4pt}
\setlength{\belowdisplayskip}{4pt}
\max\left\Vert \mathbf{v}_{\text{c}}\right\Vert \leq{v_{\text{m}}.}\label{limit}
\end{equation}
\begin{figure}[h]
\begin{centering}
\includegraphics[scale=0.9]{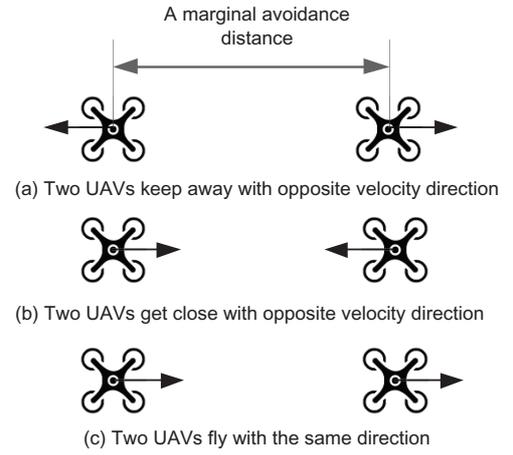} 
\par\end{centering}
\caption{Intuitive interpretation for filtered position}
\label{Intuitive}
\end{figure}

\begin{figure}[h]
\begin{centering}
\includegraphics[scale=0.75]{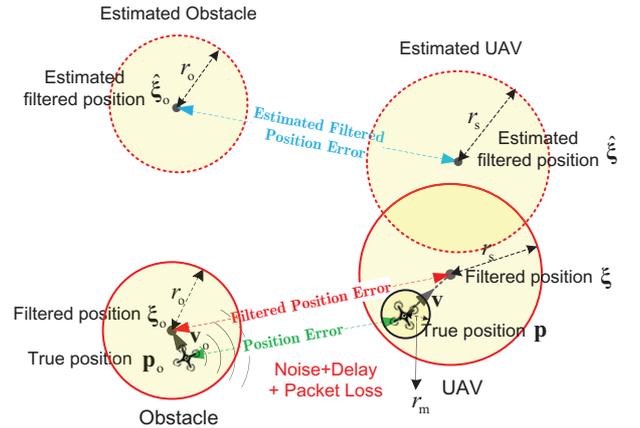} 
\par\end{centering}
\caption{Relation among position, filtered position, estimated filtered position,
where the estimated filtered position will appear in (\ref{differentialequation})
and (\ref{filteredpositionestimate}).}
\label{Filteredposition}
\end{figure}

As shown in Figure \ref{Intuitive}, although the distances between
two UAVs in the three cases are the same, namely a marginal avoidance
distance, the case in Figure \ref{Intuitive}(b) needs to carry out
avoidance urgently by considering the velocity. However, the case
in Figure \ref{Intuitive}(a) in fact does not need to be considered
to perform collision avoidance. With such an intuition, a filtered
position is defined as follows: 
\begin{equation}
\boldsymbol{\xi}\triangleq{\mathbf{p}}+\frac{1}{l}\mathbf{v}.\label{FilteredPosition}
\end{equation}
{Then} 
\begin{align}\setlength{\abovedisplayskip}{4pt}
\setlength{\belowdisplayskip}{4pt}
\boldsymbol{\dot{\xi}} & =\mathbf{\dot{p}}+\frac{1}{l}\mathbf{\dot{v}}\nonumber \\
 & =\mathbf{v}-\frac{1}{l}l\left(\mathbf{v}-\mathbf{v}_{\text{c}}\right)\nonumber \\
 & =\mathbf{v}_{\text{c}}.\label{omodel}
\end{align}
The relationship among the position, the filtered position and the
estimated filtered position is shown in Figure \ref{Filteredposition},
where the UAV cannot access the position and the filtered position
(ground truth) but only the estimated filtered position including
uncertainties. From the UAV's view, it senses itself and obstacle
at both estimated filtered positions (dash circle) in Figure \ref{Filteredposition}.
If the UAV is at  a high speed or the maneuver constant $l$ is small,
then true position ${\mathbf{p}}$ will be far from its filtered position.

\textbf{Remark 1}. It should be noted that the model (\ref{positionmodel_ab_con_i})
is a second-order system. By the transformation (\ref{FilteredPosition}),
the model (\ref{positionmodel_ab_con_i}) is transformed into a single
integrator form to simplify the further controller design and analysis.
What is more, the proposed model (\ref{positionmodel_ab_con_i}) has
taken the maneuverability into consideration. Although a commonly-used
model 
\begin{equation}
\mathbf{\dot{p}}=\mathbf{v}_{\text{c}}\label{commonmodel}
\end{equation}
is also a single integrator model, it does not take the maneuverability
into consideration. As a result, the true velocity $\mathbf{v}$ will
be different from $\mathbf{v}_{\text{c}}$. The difference will be
big, if the UAV's maneuverability is low. This further increases the
difference between the true position and desired position especially
in some situations such as making an aggressive maneuver.

\subsection{Obstacle Model}

In the same local airspace, there exists a moving obstacle (it may
be an aircraft or a balloon) defined as 
\[
\mathcal{O}=\left\{ \mathbf{x}\in\mathbb{R}^{3}\left\vert \left\Vert \mathbf{x}-\mathbf{p}_{\text{o}}\right\Vert <r_{\text{o}}\right.\right\} 
\]
where $r_{\text{o}}>0$ is the obstacle radius, and $\mathbf{p}_{\text{o}}\in\mathbb{R}^{3}$
is the center of mass of the obstacle. Define 
\[
\boldsymbol{\xi}_{\text{o}}\triangleq\mathbf{p}_{\text{o}}+\frac{1}{l}\mathbf{v}_{\text{o}}
\]
where $\mathbf{v}_{\text{o}}\in\mathbb{R}^{3}$ is the velocity of
the obstacle. The obstacle satisfies the following model
\[
\max\left\Vert \boldsymbol{\dot{\xi}}_{\text{o}}\right\Vert \leq v_{\text{o}}
\]
where $v_{\text{o}}>0.$ This is a general model for any obstacle
with bounded velocity and acceleration. Let 
\begin{equation}
\boldsymbol{\dot{\xi}}_{\text{o}}=\mathbf{a}_{\text{o}}\label{ksi0}
\end{equation}
with $\max\left\Vert \mathbf{a}_{\text{o}}\right\Vert \leq v_{\text{o}}.$
Then (\ref{ksi0}) can be rewritten as 
\begin{align}
\mathbf{\dot{p}}_{\text{o}} & =\mathbf{v}_{\text{o}}\nonumber \\
\mathbf{\dot{v}}_{\text{o}} & =-l\left(\mathbf{v}_{\text{o}}-\mathbf{a}_{\text{o}}\right).\label{obstacle}
\end{align}
In particular, if $\left\Vert \mathbf{v}_{\text{o}}\left(0\right)\right\Vert \leq v_{\text{o}}\ $and
$\mathbf{a}_{\text{o}}=\mathbf{v}_{\text{o}}\left(0\right),$ then
the obstacle is moving with a constant velocity.

\subsection{Assumptions on Uncertainties}

\textbf{Assumption 1(}Estimate noise\textbf{)}. For the UAV, the position
estimate during the flight is $\boldsymbol{\boldsymbol{\xi}}+\boldsymbol{\varepsilon},$
where $\left\Vert \boldsymbol{\varepsilon}\right\Vert \leq b$ and
$\left\Vert \boldsymbol{\dot{\varepsilon}}\right\Vert \leq v_{b}.$

\textbf{Assumption 2(}Broadcast delay \& Packet loss\textbf{)}. The
obstacle can be surveilled and then broadcast, or it can broadcast
its information to the UAV. The interval of receiving information
for the UAV is $T_{\text{s}}>0$, while the time delay (including
the broadcast period) is $0<\tau_{\text{d}}\leq\tau_{\text{dm}}.$
Let $\theta\in\left[0,1\right]$ be the probability of packet loss,
$\theta\leq\theta_{\text{m}}$. The estimate $\boldsymbol{\hat{\xi}}{_{\text{o}}}$
is a value that the UAV gets the estimated information from the obstacle
via communication with the following model
\begin{align}
\boldsymbol{\dot{\bar{\xi}}}_{\text{o}}\left(t\right) & =-\frac{1-\theta}{\theta T_{\text{s}}}\boldsymbol{\bar{\xi}}_{\text{o}}\left(t\right)+\frac{1-\theta}{\theta T_{\text{s}}}\boldsymbol{\xi}_{\text{o}}\left(t-\tau_{\text{d}}\right)\nonumber \\
\boldsymbol{\hat{\xi}}{_{\text{o}}}\left(t\right) & =\boldsymbol{\bar{\xi}}_{\text{o}}\left(t\right)+\boldsymbol{\varepsilon}_{\text{o}},\boldsymbol{\bar{\xi}}_{\text{o}}\left(0\right)=\boldsymbol{\xi}_{\text{o}}\left(-\tau_{\text{d}}\right)\label{differentialequation}
\end{align}
where $\left\Vert \boldsymbol{\varepsilon}_{\text{o}}\right\Vert \leq b_{\text{o}}\ $and
$\left\Vert \boldsymbol{\dot{\varepsilon}}_{\text{o}}\right\Vert \leq v_{b_{\text{o}}}.$

As shown in Figure \ref{Filteredposition}, the value $\boldsymbol{\hat{\xi}}_{(\cdot)}$
represents the \emph{estimated filtered position}. There exist two
cases:
\begin{itemize}
\item Information from itself. Based on \textit{Assumption 1}, as for the
UAV itself, the filtered position estimate is 
\begin{equation}
\boldsymbol{\hat{\xi}}=\boldsymbol{\xi}+\boldsymbol{\varepsilon}\label{filteredpositionestimate}
\end{equation}
because no broadcast delay and packet loss need to be considered.
\item Information from obstacle. The UAV has to receive information from
the obstacle via communication. As shown in Figure \ref{Shared},
the UAV receives the information $\boldsymbol{\hat{\xi}}{_{\text{o}}}$,
which has changed by estimate noise, broadcast delay, and packet loss.
\begin{figure}[h]
\begin{centering}
\includegraphics[scale=0.5]{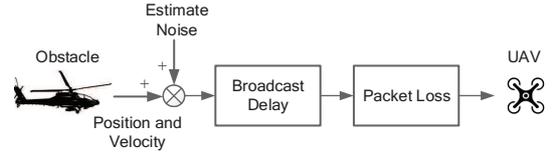} 
\par\end{centering}
\caption{Shared information broadcasting}
\label{Shared}
\end{figure}
\end{itemize}
\textbf{Remark 2}.\textbf{ }The reasonability of the selected model
(\ref{differentialequation}) is explained as follows. First, a simple
and reasonable estimation method is adopted 
\begin{equation}
\boldsymbol{\hat{\xi}}_{\text{o}}\left(t\right)=\left\{ \begin{array}{c}
\boldsymbol{\xi}_{\text{o}}\left(t-\tau_{\text{d}}\right)\\
\boldsymbol{\hat{\xi}}_{\text{o}}\left(t-T_{\text{s}}\right)
\end{array}\right.\begin{array}{l}
\text{if data packet is received}\\
\text{if data packet is lost}
\end{array}\text{.}\label{packetloss}
\end{equation}
This implies that the estimate will remain the last estimated value
if data packet is lost. Then, according to the probability of packet
loss $\theta,$ the expected value of $\boldsymbol{\hat{\xi}}_{\text{o}}$
is 
\begin{equation}
\boldsymbol{\bar{\xi}}_{\text{o}}\left(t\right)=\theta\boldsymbol{\bar{\xi}}_{\text{o}}\left(t-T_{\text{s}}\right)+\left(1-\theta\right)\boldsymbol{\xi}_{\text{o}}\left(t-\tau_{\text{d}}\right),t>0\label{averagemodel}
\end{equation}
where $\boldsymbol{\bar{\xi}}_{\text{o}}=$E$\left(\boldsymbol{\hat{\xi}}_{\text{o}}\right).$
Roughly, the differential $\boldsymbol{\bar{\xi}}_{\text{o}}$ can
be written as 
\begin{equation}
\boldsymbol{\dot{\bar{\xi}}}_{\text{o}}\left(t\right)\approx\frac{\boldsymbol{\bar{\xi}}_{\text{o}}\left(t\right)-\boldsymbol{\bar{\xi}}_{\text{o}}\left(t-T_{\text{s}}\right)}{T_{\text{s}}}.\label{eq:divapprox}
\end{equation}
Consequently, by using (\ref{eq:divapprox}), the algebraic transformation
of (\ref{averagemodel}) is 
\[
\boldsymbol{\dot{\bar{\xi}}}_{\text{o}}\left(t\right)\thickapprox-\frac{1-\theta}{\theta T_{\text{s}}}\boldsymbol{\bar{\xi}}_{\text{o}}\left(t\right)+\frac{1-\theta}{\theta T_{\text{s}}}\boldsymbol{\xi}_{\text{o}}\left(t-\tau_{\text{d}}\right).
\]
Furthermore, putting uncertainties on $\boldsymbol{\varepsilon}_{\text{o}},$
we have the differential equation model (\ref{differentialequation}).
Therefore, we can replace the model (\ref{packetloss}) with the new
model (\ref{differentialequation}). In the following simulation,
the reasonability of the selected model (\ref{differentialequation})
will be further shown.

\textbf{Remark 3}.\textbf{ }The broadcast delay $\tau_{\text{dm}}$
is the maximum delay we can accept, which is the worst case. By using
it, the most conservative safety radius is derived. If the delay exceeds
$\tau_{\text{dm}},$ then the packet can be considered as a loss.
The time delay and the probability of packet loss are very normal
parameters and can be easily measured for communication \cite{Glaab(2018)}.
For\textbf{\ }most\textbf{\ }UAVs,\textbf{ }filters will be used
to eliminate the high-frequency noise to preserve the low-frequency
information. As a result,\textbf{\ }$b,v_{b},b_{\text{o}},v_{b_{\text{o}}}$
will be small.

\subsection{Assumption on Controller}

Let 
\begin{align}
\mathbf{\tilde{p}}_{\text{o}} & \triangleq\mathbf{p}-\mathbf{p}_{\text{o}}\nonumber \\
\mathbf{\tilde{v}}_{\text{o}} & \triangleq\mathbf{v}-\mathbf{v}_{\text{o}}\nonumber \\
\boldsymbol{\tilde{\xi}}_{\text{o}} & \triangleq\boldsymbol{\xi}-\boldsymbol{\xi}_{\text{o}}.\label{errors}
\end{align}
For the UAV, no \emph{collision} with the obstacle implies 
\[
\mathcal{U}\cap\mathcal{O}=\varnothing
\]
namely \emph{true} \emph{position distance }$\left\Vert \mathbf{\tilde{p}}_{\text{o}}\right\Vert $
satisfies 
\begin{equation}
\left\Vert \mathbf{\tilde{p}}_{\text{o}}\right\Vert \geq r_{\text{m}}+r_{\text{o}}.\label{colliderequirement2}
\end{equation}

In the following, there are two assumptions about controller design.
\textit{Assumption 3 }is only for the \textit{design phase}, during
which a controller is designed for (\ref{positionmodel_ab_con_i})
without considering uncertainties to make the \emph{filtered position
distance}, namely $\left\Vert \boldsymbol{\tilde{\xi}}{}_{\text{o}}\right\Vert ,$
greater than a certain distance.

\textbf{Assumption 3 (}Controller in Design Phase\textbf{)}. Given
a \emph{designed safety radius}$\ r_{\text{s}}>0,$ with any $\boldsymbol{\tilde{\xi}}_{\text{o}}\left(0\right)\ $satisfying
$\left\Vert \boldsymbol{\tilde{\xi}}_{\text{o}}\left(0\right)\right\Vert \geq r_{\text{s}}+r_{\text{o}}{,}$\ a
controller 
\begin{equation}
\mathbf{v}_{\text{c}}=\mathbf{c}\left(t,\boldsymbol{\tilde{\xi}}_{\text{o}}\right)\label{con_designphase}
\end{equation}
for (\ref{positionmodel_ab_con_i}) can make 
\begin{equation}
\left\Vert \boldsymbol{\tilde{\xi}}_{\text{o}}\left(t\right)\right\Vert \geq r_{\text{s}}+r_{\text{o}}\label{Safetyaera}
\end{equation}
for any obstacle with $\left\Vert \boldsymbol{\dot{\xi}}_{\text{o}}\right\Vert \leq v_{\text{m}},$
where $\left\Vert \mathbf{c}\left(t,\boldsymbol{\tilde{\xi}}_{\text{o}}\right)\right\Vert \leq v_{\text{m}}$,
for $t\geq0,$ $\forall\boldsymbol{\tilde{\xi}}_{\text{o}}\in\mathbb{R}^{3}.$

\textit{Assumption 3 }is equivalent to\textit{ Lemma 1.}

\textbf{Lemma 1}. If and only if 
\begin{equation}
\underset{\mathbf{x}\in\mathcal{C}}{\min}\mathbf{x}^{\text{T}}\mathbf{c}\left(t,\mathbf{x}\right)\geq\left(r_{\text{s}}+r_{\text{o}}\right)v_{\text{m}}\label{suffnece}
\end{equation}
then (\ref{Safetyaera}) holds for any obstacle $\left\Vert \boldsymbol{\dot{\xi}}_{\text{o}}\right\Vert \leq v_{\text{m}}$
and $\left\Vert \boldsymbol{\tilde{\xi}}_{\text{o}}\left(0\right)\right\Vert \geq r_{\text{s}}+r_{\text{o}},$
where $\mathcal{C}=\left\{ \left.\mathbf{x\in}\mathbb{R}^{3}\right\vert \left\Vert \mathbf{x}\right\Vert =r_{\text{s}}+r_{\text{o}}\right\} .$

\textit{Proof}. The idea here relies on the following inequality 
\begin{equation}
\underset{\mathbf{x}\in\mathcal{C}}{\min}\mathbf{x}^{\text{T}}\mathbf{\dot{x}}\geq0.\label{L11}
\end{equation}
If and only if (\ref{L11}) holds, then $\left\Vert \mathbf{x}\left(t\right)\right\Vert \geq r_{\text{s}}+r_{\text{o}}$
for any $\mathbf{x}$ with $\left\Vert \mathbf{x}\left(0\right)\right\Vert \geq r_{\text{s}}+r_{\text{o}}.$
The remaining proofs are in \textit{Appendix}. $\square$

It is necessary to assume $\left\Vert \boldsymbol{\dot{\xi}}{_{\text{o}}}\right\Vert \leq{v_{\text{m}}}$.
Otherwise, in the worst case, the collision cannot be avoided. For
example, we can choose the obstacle dynamic as 
\begin{equation}
\boldsymbol{\dot{\xi}}{_{\text{o}}}=\boldsymbol{\dot{\xi}}-{\epsilon}\frac{\boldsymbol{\xi}{_{\text{o}}}-\boldsymbol{\xi}}{\left\Vert \boldsymbol{\xi}{_{\text{o}}}-\boldsymbol{\xi}\right\Vert }\label{Remark4}
\end{equation}
where $\left\Vert \boldsymbol{\dot{\xi}}{_{\text{o}}}\right\Vert \leq\left\Vert \boldsymbol{\dot{\xi}}\right\Vert +{\epsilon\leq v_{\text{m}}+\epsilon}$
{with} ${{\epsilon}>0.}$ From (\ref{Remark4}), it is easy to see
$\left\Vert \boldsymbol{\xi}_{\text{o}}\left(t\right)-\boldsymbol{\xi}\left(t\right)\right\Vert <r_{\text{s}}+r_{\text{o}}$
within a finite time no matter how small ${\epsilon}$ is. This implies
that the obstacle is chasing after the UAV and then hits it finally.

In \textit{Assumption 3}, we do not consider the uncertainties because
the communication uncertainties will be left to the designed safety
radius so that the same controller with different design safety radiuses
can deal with different communication uncertainties. In the practice
flight phase, $\mathbf{e}_{\text{o}}$ rather than $\boldsymbol{\tilde{\xi}}{_{\text{o}}}$\ is
only available. So, the same controller is used but with different
feedback such as 
\begin{equation}
\mathbf{v}_{\text{c}}=\mathbf{c}\left(t,\mathbf{e}_{\text{o}}\right)\label{con_flightphase}
\end{equation}
for (\ref{positionmodel_ab_con_i}), where 
\begin{equation}
\mathbf{e}_{\text{o}}\triangleq\boldsymbol{\hat{\xi}}-\boldsymbol{\hat{\xi}}_{\text{o}}.\label{eo1}
\end{equation}
Since the controller cannot get the truth, we only can make the \emph{estimated
filtered position distance}, namely $\left\Vert \mathbf{e}_{\text{o}}\right\Vert ,$
greater than a certain distance. \textit{Assumption 4 }is only for
the \textit{flight phase} in the presence of uncertainties, describing
what has happened.

\textbf{Assumption 4 (}Control Performance in Flight Phase\textbf{)}.
There exists a \emph{practical safety radius}$\ r_{\text{s}}^{\prime}>0\ $such
that, for any obstacle with $\left\Vert \boldsymbol{\dot{\xi}}{_{\text{o}}}\right\Vert \leq{v_{\text{m}}\ }${and}$\ \left\Vert \mathbf{e}_{\text{o}}\left(0\right)\right\Vert \geq r_{\text{s}}^{\prime}+r_{\text{o}},$
a controller for (\ref{positionmodel_ab_con_i}) can make 
\begin{equation}
\left\Vert \mathbf{e}_{\text{o}}\left(t\right)\right\Vert \geq r_{\text{s}}^{\prime}+r_{\text{o}}\label{Safetyaera1}
\end{equation}
where $t\geq0.$

\subsection{Objective}

In the `offline' design phase (in numerical simulation), a question
will arise that \emph{how far} should the UAV and the obstacle be
kept without uncertainties offline in order to avoid a collision in
the presence of uncertainties?
\begin{itemize}
\item \textbf{Design Phase Objective (Offline)}. Under \textit{Assumptions
1-3 }and controller\textit{\ }(\ref{con_designphase}) for (\ref{positionmodel_ab_con_i}),
the objective here is to determine the \emph{estimated safety radius}
$\hat{r}_{\text{s}}>0$ (it is related to $r_{\text{s}}$) that 
\begin{equation}
\left\Vert \mathbf{e}_{\text{o}}\left(t\right)\right\Vert \geq\hat{r}_{\text{s}}+r_{\text{o}}\label{eo}
\end{equation}
holds, where $t\geq0.$ Furthermore, determine the designed safety
radius $r_{\text{s}}$ to make\textit{\ }(\ref{colliderequirement2})
hold\ in the presence of uncertainties. 
\end{itemize}
On the other hand, when the UAV in practice (in the presence of uncertainties),
a question will arise that \emph{how far} should the UAV and the obstacle
be kept in the sense of the estimated filtered position distance,
namely $\left\Vert \mathbf{e}_{\text{o}}\right\Vert ,$ in order to
avoid a collision?
\begin{itemize}
\item \textbf{Flight Phase Objective (Online)}. Under \textit{Assumptions
1-2,4},\textit{\ }the second objective is to determine the \emph{practical
safety radius} $r_{\text{s}}^{\prime}>0$ to satisfy (\ref{colliderequirement2}). 
\end{itemize}
\textbf{Remark 4}. The design phase objective is a type of the principle
of separation of control and safety radius. It can be stated that
\emph{under some assumptions, the problem of designing a collision-avoidance
controller with uncertainties can be solved by designing a safety
radius covering the uncertainties, which feeds into a deterministic
collision avoidance controller for the system. }Thus, the problem
can be broken into two separate parts, which facilitates the design.
One controller with different designed safety radiuses can deal with
different communication uncertainties. In some cases, controllers
may not satisfy \textit{Assumption 3} but \textit{Assumption 4 }in
practice. This is because the controller in \textit{Assumption 3 }considers
any initial condition, or say the worst condition. But it is not necessary.
For example, a UAV and an obstacle are supposed to only fly along
two separate airlines without any collision avoidance control. In
this case, \textit{Assumption 3 }is not satisfied, but \textit{Assumption
4 }may hold. This motivates us to consider the safety radius separately
in\emph{ }the design phase and the flight phase.

\section{Safety Radius Design}

Preliminaries are given first. By using them, the estimated safety
radius and the practical safety radius are designed for the UAV and
one obstacle. Furthermore, the results are extended to multiple obstacles.

\subsection{Preliminary}

First, an important lemma is proposed.

\textbf{Lemma 2}.\textbf{\ }Let $\mathbf{x}\left(t\right)\in{{\mathbb{R}}^{n}}$
and $\mathbf{y}\left(t\right)\in{{\mathbb{R}}^{n}}$ satisfy 
\begin{equation}
\mathbf{\dot{x}}\left(t\right)=-k\left(t\right)\mathbf{x}\left(t\right)+k\left(t\right)\mathbf{y}\left(t\right)\label{equlemma1}
\end{equation}
where $0<k_{\min}\leq k\left(t\right)\leq k_{\max}.$ If $\left\Vert \mathbf{y}\left(t\right)\right\Vert \leq y_{\max},$
$\left\Vert \mathbf{\dot{y}}\left(t\right)\right\Vert \leq v_{y_{\max}},$
and $\left\Vert \mathbf{x}\left(0\right)\right\Vert \leq y_{\max}$,
then 
\begin{equation}
\left\Vert \mathbf{x}\left(t\right)\right\Vert \leq y_{\max},t\geq0.\label{conlemma11}
\end{equation}
If $\left\Vert \mathbf{x}\left(0\right)-\mathbf{y}\left(0\right)\right\Vert \leq\frac{1}{k_{\min}}v_{y_{\max}}$
holds, then 
\begin{equation}
\left\Vert \mathbf{\dot{x}}\left(t\right)\right\Vert \leq\frac{k_{\max}}{k_{\min}}v_{y_{\max}},t\geq0.\label{conlemma12}
\end{equation}

\textit{Proof}. See \textit{Appendix}. $\square$

With \textit{Lemma 2 }in hand, we have

\textbf{Proposition 1}. If $\left\Vert \mathbf{v}\left(0\right)\right\Vert \leq{v_{\text{m}}}$
and the model (\ref{positionmodel_ab_con_i}) is subject to (\ref{limit}),
then $\left\Vert \mathbf{v}\left(t\right)\right\Vert \leq{v_{\text{m}},}$
$t\geq0{.}$

\textit{Proof}. It is easy from \textit{Lemma 2}. $\square$

According to \textit{Proposition 1}, we have 
\begin{equation}
\frac{1}{{l}}\left\Vert \mathbf{\tilde{v}}{_{\text{o}}}\right\Vert \leq r_{\text{v}}\label{rv1}
\end{equation}
where 
\begin{equation}
r_{\text{v}}=\frac{{v_{\text{m}}}+v_{\text{o}}}{{l}}.\label{rv}
\end{equation}
In the following, a relationship between the true position error $\mathbf{\tilde{p}}{_{\text{o}}}$
and the filtered position error $\boldsymbol{\tilde{\xi}}{_{\text{o}}}$\ is
shown. \textit{Proposition 2} implies that the UAV and the obstacle
will be separated largely enough if their filtered position distance
is separated largely enough.

\textbf{Proposition 2}. For the VTOL UAV and the obstacle, \emph{if
and only if} the filtered position error satisfies 
\begin{equation}
\left\Vert \boldsymbol{\tilde{\xi}}{_{\text{o}}}\left(t\right)\right\Vert \geq\sqrt{r^{2}+r_{\text{v}}^{2}}\label{p3condition}
\end{equation}
and $\left\Vert \mathbf{\tilde{p}}{_{\text{o}}}\left(0\right)\right\Vert \geq r,$
then $\left\Vert \mathbf{\tilde{p}}{_{\text{o}}}\left(t\right)\right\Vert \geq r,$
where $t>0$. The relationship ``$=$''\ holds if $\frac{\mathbf{v}^{\text{T}}{{\mathbf{v}}_{\text{o}}}}{\left\Vert \mathbf{v}\right\Vert \left\Vert {{\mathbf{v}}_{\text{o}}}\right\Vert }=-1.$

\textit{Proof}. See\textit{\ Appendix. }$\square$

\subsection{Separation Principle}

For the design phase, the principle of separation of controller is
stated in \textit{Theorem 1}.

\textbf{Theorem 1 (}Separation Theorem\textbf{)}.\textbf{ }Suppose
that\textbf{ }the UAV is with model (\ref{positionmodel_ab_con_i})
under \textit{Assumptions 1-2}. Then (i)\textit{ }\emph{if and only
if} 
\begin{equation}
\left.\left(\mathbf{e}_{\text{o}}^{\text{T}}\boldsymbol{\dot{\xi}}-\mathbf{e}_{\text{o}}^{\text{T}}\boldsymbol{\dot{\hat{\xi}}}{_{\text{o}}}\right)\right\vert _{\left\Vert \mathbf{e}_{\text{o}}\right\Vert =r_{\text{s}}+r_{\text{o}}}\geq\left(r_{\text{s}}+r_{\text{o}}\right)v_{b}\label{ConTheorem1}
\end{equation}
then (\ref{eo}) holds with $\hat{r}_{\text{s}}=r_{\text{s}}\ $for
any $\left\Vert \mathbf{e}_{\text{o}}\left(0\right)\right\Vert \geq r_{\text{s}}+r_{\text{o}}$;
(ii) furthermore, under \textit{Assumptions 1-3},\textit{ }if 
\begin{equation}
\left(r_{\text{s}}+r_{\text{o}}\right)\left(v_{\text{m}}-v_{b}\right)\geq\left.\mathbf{e}_{\text{o}}^{\text{T}}\boldsymbol{\dot{\hat{\xi}}}{_{\text{o}}}\right\vert _{\left\Vert \mathbf{e}_{\text{o}}\right\Vert =r_{\text{s}}+r_{\text{o}}}\label{cooperconditionTh1}
\end{equation}
then (\ref{eo}) holds with $\hat{r}_{\text{s}}=r_{\text{s}}$; (iii)
in particular, under \textit{Assumptions 1-3},\textit{ }if 
\begin{equation}
v_{\text{m}}\geq{v_{\text{o}}}+v_{b}+v_{b_{\text{o}}},\label{noncooperconditionTh1}
\end{equation}
\textit{\ }then (\ref{eo}) holds with $\hat{r}_{\text{s}}=r_{\text{s}}.$

\textit{Proof}. See\textit{\ Appendix. }$\square$
\begin{figure}[h]
\begin{centering}
\includegraphics[scale=0.7]{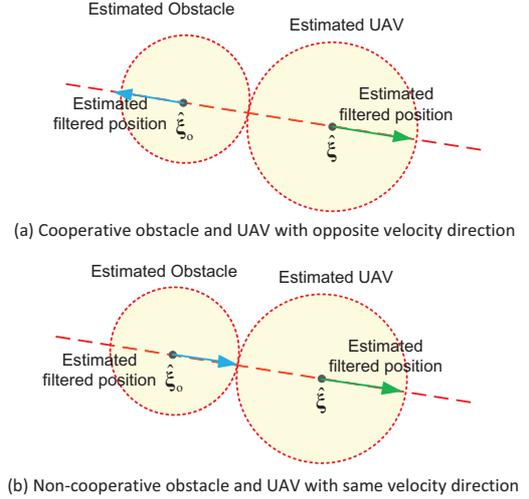} 
\par\end{centering}
\caption{A cooperative obstacle and a non-cooperative obstacle}
\label{seperation}
\end{figure}

\textbf{Remark 5}.\textbf{ }Through \emph{Theorem 1}, we obtain $\left\Vert \mathbf{e}_{\text{o}}\left(t\right)\right\Vert \geq\hat{r}_{\text{s}}+r_{\text{o}}$,
based on which we can directly determine the lower bound of the safety
radius to make UAVs safe. From now on, we do not need to consider
the controller any more, because \emph{Theorem 1 }has \emph{seperated}
the controller design and safety radius design. So, we call \emph{Theorem
1} as \emph{separation theorem.}

\textbf{Remark 6}.\textbf{ }Let us consider a cooperative obstacle
in the best case shown in Figure \ref{seperation}(a) and a non-cooperative
obstacle in the worst case shown in Figure \ref{seperation}(b). The
cooperative obstacle can be considered as another UAV with the same
controller, namely 
\[
\mathbf{a}_{\text{o}}=\mathbf{c}\left(t,-\mathbf{e}_{\text{o}}\right)
\]
where, for simplicity, $-\mathbf{e}_{\text{o}}$ is used for another
UAV's feedback approximately. The obstacle (another UAV) will, in
turn, take the UAV as its ``obstacle''\ and will make collision
avoidance simultaneously as well. According to \textit{Lemma 1},\textbf{
}the following inequality 
\[
\left.-\mathbf{e}_{\text{o}}^{\text{T}}\mathbf{c}\left(t,-\mathbf{e}_{\text{o}}\right)\right\vert _{\left\Vert \mathbf{e}_{\text{o}}\right\Vert =r_{\text{s}}+r_{\text{o}}}\geq\left(r_{\text{s}}+r_{\text{o}}\right){v_{\text{m}}}
\]
still holds. Since $\boldsymbol{\dot{\xi}}{_{\text{o}}}=\mathbf{c}\left(t,-\mathbf{e}_{\text{o}}\right),$
we have 
\[
\left.\mathbf{e}_{\text{o}}^{\text{T}}\boldsymbol{\dot{\xi}}{_{\text{o}}}\right\vert _{\left\Vert \mathbf{e}_{\text{o}}\right\Vert =r_{\text{s}}+r_{\text{o}}}\leq-\left(r_{\text{s}}+r_{\text{o}}\right){v_{\text{m}}<0.}
\]
If $\boldsymbol{\dot{\hat{\xi}}}{_{\text{o}}}\approx\boldsymbol{\dot{\xi}}{_{\text{o}},}$
then $\left.\mathbf{e}_{\text{o}}^{\text{T}}\boldsymbol{\dot{\hat{\xi}}}{_{\text{o}}}\right\vert _{\left\Vert \mathbf{e}_{\text{o}}\right\Vert =r_{\text{s}}+r_{\text{o}}}{<0.}$
Therefore, (\ref{cooperconditionTh1}) is satisfied in most cases
without the requirement (\ref{noncooperconditionTh1}). Figure \ref{seperation}(a)
shows the best case that the UAV and obstacle can keep away simultaneously
with the opposite direction. Intuitively, no limitation will be put
on $v_{\text{o}}.$ Figure \ref{seperation}(b) shows the worst case
that the UAV and obstacle move simultaneously in the same direction.
In this case, the requirement (\ref{noncooperconditionTh1}) implies
that the UAV should have a faster speed than the obstacle's speed.

\subsection{Safety Radius Design}

\begin{figure}[h]
\begin{centering}
\includegraphics[scale=0.7]{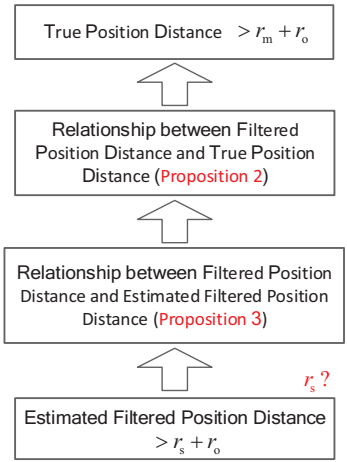} 
\par\end{centering}
\caption{Relationship from true position distance to estimated filtered position
distance}
\label{turetoestimatedfiltered}
\end{figure}

Based on $r_{\text{m}},r_{\text{o}}>0$, we will further determine
$r_{\text{s}}>0\ $in order to avoid a collision in the presence of
uncertainties. For this purpose, we need to analyze the relationship
between the filtered position distance and the true position distance
(\textit{Proposition 2 }has done), and the relationship between the
filtered position distance and the estimated filtered position distance
(\textit{Proposition 3 }will show), as shown in Figure \ref{turetoestimatedfiltered}.

\textbf{Proposition 3}. Under \textit{Assumptions 1-2}, given any
$r>0,$ if 
\begin{equation}
\left\Vert \mathbf{e}_{\text{o}}\left(t\right)\right\Vert \geq r+r_{e},t\geq0\label{p3_eo}
\end{equation}
then 
\begin{equation}
\left\Vert \boldsymbol{\tilde{\xi}}_{\text{o}}\left(t\right)\right\Vert \geq r,t\geq0\label{ksi_eo}
\end{equation}
where 
\begin{equation}
r_{e}=\frac{\theta_{\text{m}}T_{\text{s}}}{1-\theta_{\text{m}}}v_{\text{o}}+v_{\text{o}}\tau_{\text{dm}}+b+b_{\text{o}}\label{rji}
\end{equation}

\textit{Proof}. See\textit{\ Appendix. }$\square$

With \textit{Lemma 2}, we start to determine the lower bound of the
safety radius $r_{\text{s}}\ $in the design phase.

\textbf{Theorem 2}. Under \textit{Assumptions 1-3}, if the designed
safety radius satisfies 
\begin{equation}
r_{\text{s}}\geq\sqrt{\left(r_{\text{m}}+r_{\text{o}}\right)^{2}+r_{\text{v}}^{2}}+r_{e}-r_{\text{o}},\label{Th11}
\end{equation}
then condition (\ref{colliderequirement2}) holds.

\textit{Proof.} According to \textit{Proposition 3},\textbf{ }if 
\[
\left\Vert \mathbf{e}_{\text{o}}\left(t\right)\right\Vert \geq r_{\text{s}}+r_{\text{o}}
\]
then 
\[
\left\Vert \boldsymbol{\tilde{\xi}}_{\text{o}}\left(t\right)\right\Vert \geq r_{\text{s}}+r_{\text{o}}-r_{e}.
\]
If (\ref{Th11}) holds, then 
\[
\left(r_{\text{s}}+r_{\text{o}}-r_{e}\right)^{2}\geq\left(r_{\text{m}}+r_{\text{o}}\right)^{2}+r_{\text{v}}^{2}
\]
namely 
\[
\left\Vert \boldsymbol{\tilde{\xi}}{_{\text{o}}}\left(t\right)\right\Vert \geq\sqrt{\left(r_{\text{m}}+r_{\text{o}}\right)^{2}+r_{\text{v}}^{2}}.
\]
According to \textit{Proposition 2},\textit{\ }we have 
\[
\left\Vert \mathbf{\tilde{p}}{_{\text{o}}}\left(t\right)\right\Vert \geq r_{\text{m}}+r_{\text{o}}.
\]
$\square$

With \textit{Theorem 2} in hand, the solution to the flight phase
objective is easy to get.

\textbf{Theorem 3}. Under \textit{Assumptions 1-2,4}, if the practical
safety radius satisfies 
\[
r_{\text{s}}^{\prime}\geq\sqrt{\left(r_{\text{m}}+r_{\text{o}}\right)^{2}+r_{\text{v}}^{2}}+r_{e}-r_{\text{o}}
\]
then (\ref{colliderequirement2}) holds.

\textit{Proof.} It is similar to the proof of \textit{Theorem 2}.
$\square$

\subsection{Extension to Multiple Obstacles}

The results above can be extended to multiple obstacles as well. There
are ${M}$ obstacles 
\[
\mathcal{O}_{\text{o,}k}=\left\{ \mathbf{x}\in{{\mathbb{R}}^{3}}\left\vert \left\Vert \mathbf{x}-{{\mathbf{p}}_{\text{o,}k}}\right\Vert \leq{{r}_{\text{o}}}\right.\right\} 
\]
where ${{\mathbf{p}}_{\text{o,}k}}\in{{\mathbb{R}}^{3}}$ is the center
position of the $k$th obstacle, $\mathbf{v}{_{\text{o,}k}={\mathbf{\dot{p}}}_{\text{o,}k}}\in{{\mathbb{R}}^{3}}$
is the velocity of the $k$th obstacle, $k=1,\cdots,{M}$. Define
\begin{align*}
\boldsymbol{\xi}_{\text{o,}k} & \triangleq\mathbf{p}_{\text{o,}k}+\frac{1}{{l}}\mathbf{v}_{\text{o,}k}\\
\boldsymbol{\tilde{\xi}}_{\text{o,}k} & \triangleq\boldsymbol{\xi}-\boldsymbol{\xi}_{\text{o,}k}.
\end{align*}
These obstacles satisfy $\max\left\Vert \boldsymbol{\dot{\xi}}_{\text{o,}k}\right\Vert \leq v_{\text{o}}$,
$k=1,\cdots,{M}$. For the UAV, no \emph{collision} with the multiple
obstacles implies 
\begin{equation}
\mathcal{\mathcal{U}}\cap\mathcal{O}_{\text{o,}k}=\varnothing\label{nocollision}
\end{equation}
where $k=1,\cdots,{M}.$ To extend the conclusions in \textit{Theorems
1-2 }to\textit{\ }multiple moving obstacles, we have \textit{Assumptions
2',3'} to replace with \textit{Assumptions 2,3 }in the following.

\textbf{Assumption 2' (}Broadcast delay \& Packet loss\textbf{)}.
The $k$th obstacle can be surveilled and then broadcast, or it can
broadcast its information to the UAV. The interval of receiving information
for the UAV is $T_{\text{s}}>0$, while the time delay (including
the broadcast period) of the $k$th obstacle is $0<\tau_{\text{d,}k}\leq\tau_{\text{dm}}.$
Let $\theta_{k}\in\left[0,1\right]$ be the probability of packet
loss for\ the $k$th obstacle,$\ \theta_{k}\leq\theta_{\text{m}}<1$\ .
The estimate $\boldsymbol{\hat{\xi}}_{\text{o},k}$ is a value that
the $k$th UAV gets the estimated information from the obstacle via
communication with the following model 
\begin{align}
\boldsymbol{\dot{\bar{\xi}}}_{\text{o},k}\left(t\right) & =-\frac{1-\theta_{k}}{\theta_{k}T_{\text{s}}}\boldsymbol{\bar{\xi}}_{\text{o},k}\left(t\right)+\frac{1-\theta_{k}}{\theta_{k}T_{\text{s}}}\boldsymbol{\xi}_{\text{o},k}\left(t-\tau_{\text{d,}k}\right)\nonumber \\
\boldsymbol{\hat{\xi}}_{\text{o},k}\left(t\right) & =\boldsymbol{\bar{\xi}}_{\text{o},k}\left(t\right)+\boldsymbol{\varepsilon}_{\text{o},k},\boldsymbol{\bar{\xi}}_{\text{o},k}\left(0\right)=\boldsymbol{\xi}_{\text{o},k}\left(-\tau_{\text{d,}k}\right)\label{differentialequationk}
\end{align}
where $\left\Vert \boldsymbol{\varepsilon}_{\text{o},k}\right\Vert \leq b_{\text{o}}\ $and
$\left\Vert \boldsymbol{\dot{\varepsilon}}_{\text{o},k}\right\Vert \leq v_{b_{\text{o}}},$
$k=1,\cdots,{M.}$
\begin{figure}[h]
\begin{centering}
\includegraphics[scale=0.5]{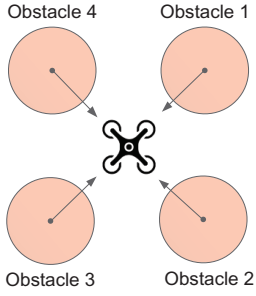} 
\par\end{centering}
\caption{UAV surrounded by four obstacles}
\label{surround}
\end{figure}

The avoidance case with multiple obstacles is complex. Under some
initial conditions, the UAV cannot avoid collision with obstacles
no matter what a controller uses, such as a case shown in Figure \ref{surround}.
For such a purpose, we define a set $\mathcal{F}$ for multiple obstacles'
and UAV's initial conditions in \textit{Assumption 3'}.

\textbf{Assumption 3'}. Given a \emph{designed safety radius}$\ r_{\text{s}}>0,$
with any $\left(\boldsymbol{\tilde{\xi}}_{\text{o,}1}\left(0\right),\cdots,\boldsymbol{\tilde{\xi}}_{\text{o,}M}\left(0\right)\right)\in\mathcal{F}{,}$\ a
controller 
\begin{equation}
\mathbf{v}_{\text{c}}=\mathbf{c}\left(t,\boldsymbol{\tilde{\xi}}{_{\text{o,}1},\cdots,}\boldsymbol{\tilde{\xi}}{_{\text{o,}M}}\right)\label{positionmodel_ab_con_k}
\end{equation}
for (\ref{positionmodel_ab_con_i}) can make 
\begin{equation}
\left\Vert \boldsymbol{\tilde{\xi}}_{\text{o,}k}\left(t\right)\right\Vert \geq r_{\text{s}}+r_{\text{o}}\label{Safetyaerak}
\end{equation}
for obstacles with $\left\Vert \boldsymbol{\dot{\xi}}{_{\text{o},k}}\right\Vert \leq{v}_{\text{o}}{,}$
where $\left\Vert \mathbf{c}\left(t,\boldsymbol{\tilde{\xi}}{_{\text{o,}1},\cdots,}\boldsymbol{\tilde{\xi}}{_{\text{o,}M}}\right)\right\Vert \leq{v_{\text{m}}}$,
for $t\geq0,$ $k=1,\cdots,{M}.$

\textbf{Theorem 4}.\textbf{ }Suppose that\textbf{ }the UAV is with
model (\ref{positionmodel_ab_con_i}) under \textit{Assumptions 1-2}.
Then (i) if and only if 
\begin{equation}
\left.\left(\mathbf{e}_{\text{o,}k}^{\text{T}}\boldsymbol{\dot{\xi}}-\mathbf{e}_{\text{\text{o,}}k}^{\text{T}}\boldsymbol{\dot{\hat{\xi}}}{_{\text{o,}k}}\right)\right\vert _{\left\Vert \mathbf{e}_{\text{o,}k}\right\Vert =r_{\text{s}}+r_{\text{o}}}\geq\left(r_{\text{s}}+r_{\text{o}}\right)v_{b}\label{CondtionTheorem4}
\end{equation}
then 
\begin{equation}
\left\Vert \mathbf{e}_{\text{o,}k}\left(t\right)\right\Vert \geq r_{\text{s}}+r_{\text{o}}\label{eok}
\end{equation}
for any $\left(\mathbf{e}_{\text{o,}1}\left(0\right),\cdots,\mathbf{e}_{\text{o,}M}\left(0\right)\right)\in\mathcal{F}$
where $\mathbf{e}_{\text{o},k}\triangleq\boldsymbol{\hat{\xi}}-\boldsymbol{\hat{\xi}}_{\text{o},k},$
$k=1,\cdots,{M}.$ (ii) In particular, under \textit{Assumptions 1,2',3'}
,\textit{ }if (\ref{noncooperconditionTh1}) holds,\textit{\ }then
(\ref{eok}), where $k=1,\cdots,{M}.$ Furthermore, if $r_{\text{s}}$
satisfies (\ref{Th11}), then (\ref{nocollision}) holds for $k=1,\cdots,{M}$.

\textit{Proof}. Proof of Conclusion (i) does not rely on \textit{Assumption
3'},\textit{ }which is similar to \textit{Conclusion (i)} of\textit{
Theorem 1.} So, we omit it. Let us prove Conclusion (ii). The controller
(\ref{positionmodel_ab_con_k}) is rewritten as 
\begin{align*}
\mathbf{v}_{\text{c}} & =\mathbf{c}\left(t,\mathbf{e}_{\text{o,1}},\cdots,\mathbf{e}_{\text{o,}M}\right)\\
 & =\mathbf{c}\left(t,\boldsymbol{\xi}-\boldsymbol{\xi}_{\text{o,1}}^{\prime},\cdots,\boldsymbol{\xi}-\boldsymbol{\xi}_{\text{o,}M}^{\prime}\right)
\end{align*}
where $\boldsymbol{\xi}_{\text{o,}k}^{\prime}\triangleq\boldsymbol{\hat{\xi}}_{\text{o},k}-\boldsymbol{\varepsilon}.$
New obstacle$\ \mathcal{O}_{k}^{\prime}$ with filtered position $\boldsymbol{\xi}_{\text{o,}k}^{\prime}$
are taken into consideration, where $k=1,\cdots,{M}$.. If $\left(\boldsymbol{\xi}\left(0\right)-\boldsymbol{\xi}_{\text{o,1}}^{\prime}\left(0\right),\cdots,\boldsymbol{\xi}\left(0\right)-\boldsymbol{\xi}_{\text{o,}M}^{\prime}\left(0\right)\right)\in\mathcal{F}$
and $\left\Vert \boldsymbol{\dot{\xi}}_{\text{o,}k}^{\prime}\right\Vert \leq{v}_{\text{o}}{,}$
then 
\[
\left\Vert \boldsymbol{\xi}-\boldsymbol{\xi}_{\text{o,}k}^{\prime}\right\Vert \geq r_{\text{s}}+r_{\text{o}}
\]
namely (\ref{eok}) holds, according to \textit{Assumption 3'},\textit{
}$k=1,\cdots,{M}$. The left problem is to study the condition $\left\Vert \boldsymbol{\dot{\xi}}_{\text{o,}k}^{\prime}\right\Vert \leq{v_{\text{m}}.}$
The derivative $\boldsymbol{\xi}_{\text{o,}k}^{\prime}$ is 
\begin{equation}
\boldsymbol{\dot{\xi}}_{\text{o,}k}^{\prime}=\boldsymbol{\dot{\bar{\xi}}}_{\text{o,}k}+\boldsymbol{\dot{\varepsilon}}_{\text{o,}k}-\boldsymbol{\dot{\varepsilon}}.\label{newobstacle}
\end{equation}
In view of (\ref{differentialequation}), according to \emph{Lemma
1}, we have 
\begin{align*}
\left\Vert \boldsymbol{\dot{\bar{\xi}}}_{\text{o,}k}\right\Vert  & \leq\left\Vert \boldsymbol{\dot{\xi}}_{\text{o,}k}\left(t-\tau_{\text{d,}k}\right)\right\Vert \\
 & \leq{v_{\text{o}}.}
\end{align*}
Then, (\ref{newobstacle}) is bounded as 
\begin{align*}
\left\Vert \boldsymbol{\dot{\xi}}_{\text{o,}k}^{\prime}\right\Vert  & \leq\left\Vert \boldsymbol{\dot{\bar{\xi}}}_{\text{o,}k}\right\Vert +\left\Vert \boldsymbol{\dot{\varepsilon}}_{\text{o,}k}\right\Vert +\left\Vert \boldsymbol{\dot{\varepsilon}}\right\Vert \\
 & \leq{v_{\text{o}}}+v_{b_{\text{o}}}+v_{b}
\end{align*}
where \textit{Assumptions 1,2'} are utilized. Therefore, if (\ref{noncooperconditionTh1})
holds, then (\ref{eok}) holds with $\hat{r}_{\text{s}}=r_{\text{s}}.$
$\square$

\textbf{Remark 7}.\textbf{ }The introduction to the set $\mathcal{F}$
is to make the problem completed, which is out of the scope of this
paper. How to find the set $\mathcal{F}$ is an interesting problem,
which can be formulated as: given a $T>0,$ the initial condition
set $\left(\boldsymbol{\tilde{\xi}}_{\text{o,}1}\left(0\right),\cdots,\boldsymbol{\tilde{\xi}}_{\text{o,}M}\left(0\right)\right)\in\mathcal{F}$
is a set that can make 
\[
\begin{tabular}{l}
 \ensuremath{\underset{\mathbf{v}_{\text{c}}}{\max}\underset{\mathbf{a}_{\text{o,}1},\cdots\mathbf{a}_{\text{o,}M}}{\min}\left(\left\Vert \boldsymbol{\tilde{\xi}}{_{\text{o,}1}}\left(t\right)\right\Vert {,\cdots,}\left\Vert \boldsymbol{\tilde{\xi}}{_{\text{o,}M}}\left(t\right)\right\Vert \right)\geq r_{\text{s}}+r_{\text{o}},0\leq t\leq T}\\
 \ensuremath{\text{s.t. }\boldsymbol{\dot{\tilde{\xi}}}{_{\text{o,}k}}=\mathbf{v}_{\text{c}}-\mathbf{a}_{\text{o,}k},\text{ }\left\Vert \mathbf{v}_{\text{c}}\right\Vert \leq v_{\text{m}},\left\Vert \mathbf{a}_{\text{o,}k}\right\Vert \leq{v}_{\text{o}},k=1,\cdots,M.} 
\end{tabular}\ 
\]
Interested readers can take the problem as the feasibility of the
pursuit-evasion game problem with multiple entities chasing a single
target or prey \cite{Vidal(2002)} or group chase and escape problem
\cite{Atsushi(2010)}.

To extend the conclusions in \textit{Theorem 3 }to\textit{\ }multiple
moving obstacles, we have \textit{Assumption 4'} to replace with \textit{Assumption
4 }in the following.

\textbf{Assumption 4'}. There exists a practical safety radius$\ r_{\text{s}}^{\prime}>0\ $such
that, for obstacles with $\left\Vert \boldsymbol{\dot{\xi}}{_{\text{o},k}}\right\Vert \leq{v}_{\text{o}}{\ }${and}$\ \left\Vert \mathbf{e}_{\text{o},k}\left(0\right)\right\Vert \geq r_{\text{s}}^{\prime}+r_{\text{o}},$
a controller for (\ref{positionmodel_ab_con_i}) can make 
\[
\left\Vert \mathbf{e}_{\text{o},k}\left(0\right)\right\Vert \geq r_{\text{s}}^{\prime}+r_{\text{o}}
\]
where $k=1,\cdots,{M}$.

In the flight phase, the results in \textit{Theorem 3 }are extended
in the following \textit{Theorem 5} for multiple obstacles.

\textbf{Theorem 5}. Under \textit{Assumptions 1,2',4'}, if the practical
safety radius satisfies 
\[
r_{\text{s}}^{\prime}\geq\sqrt{\left(r_{\text{m}}+r_{\text{o}}\right)^{2}+r_{\text{v}}^{2}}+r_{e}-r_{\text{o}}
\]
then (\ref{nocollision}) holds for $k=1,\cdots,{M}$.

\textit{Proof.} It is similar to the proof of \textit{Theorem 3}.
$\square$

\section{Simulation and Experiments}

\subsection{Simulation}

In the first two simulations for non-cooperative obstacles,\ with
the proposed separation principle (\textit{Theorems 1,4}) and the
designed safety radius (\textit{Theorem 2}), we will show that the
condition (\ref{noncooperconditionTh1}) is necessary in order to
make avoidance subject to uncertainties. However, for cooperative
obstacles which can make avoidance simultaneously, the condition (\ref{noncooperconditionTh1})
is not necessary, which is shown in the last simulation. The results
of \textit{Theorems 3,5 }can be observed directly from these results
of the following simulations by choosing $\hat{r}_{\text{s}}=r_{\text{s}}.$
A video about simulations and experiments is available on https://youtu.be/MawyB3eoZQ0
or http://t.cn/A6ZD7otD.

\subsubsection{Simulation with One Non-Cooperative Obstacle}
\begin{itemize}
\item \textbf{Simulation Setting}. As shown in Figure \ref{flight}, a scenario
that one static UAV makes avoidance with one moving non-cooperative
obstacle is considered. The simulation parameters are set as follows.
The UAV with a physical radius $r_{\text{m}}$ $=5$m is at $\mathbf{p}\left(0\right)=\left[0~0\text{ }100\right]^{\text{T}}$m
initially. The UAV's maneuver constant is $l=5$, and the minimum
speed $v_{\text{m}}=10\text{m/s}$. The obstacle is at $\mathbf{p}_{\text{o}}\left(0\right)=\left[40~0\text{ }100\right]^{\text{T}}$m
initially with radius $r_{\text{o}}=10$m and a constant velocity
$\mathbf{v}_{\text{o}}=\left[-5~0~0\right]^{\text{T}}$m/s. The interval
of receiving information for the UAV is $T_{\text{s}}=0.01$s. Communication
uncertainty parameters are set as Table I, where only \textit{Case
A} has no uncertainties. We make \textit{Case B} satisfy the condition
of (\ref{noncooperconditionTh1})\ but \textit{Case C}\ not intentionally.
The designed safety radiuses are all chosen according to (\ref{Th11})
in \textit{Theorem 2.} 
\begin{table}[ptb]
\begin{centering}
\begin{tabular}{|c|c|c|c|c|c|c|c|}
\hline 
Case {*}  & $b$(m)  & $b_{\text{o}}$(m)  & $v_{b}$(m/s)  & $v_{b_{\text{o}}}$(m/s)  & $\tau_{\text{d}}$(s)  & $\theta$  & $r_{\text{s}}$(m)\tabularnewline
\hline 
Case A  & 0  & 0  & 0  & 0  & 0  & 0  & 5.30\tabularnewline
\hline 
Case B  & 3  & 1  & 3  & 1  & 1  & 10\%  & 14.30\tabularnewline
\hline 
Case C  & 5  & 2  & 6  & 5  & 2  & 20\%  & 22.31\tabularnewline
\hline 
\end{tabular}
\par\end{centering}
\caption{Different communication parameters}
\end{table}
\item \textbf{Assumption Verification}. The comparison of model (\ref{differentialequation})
in \textit{Assumption 2 }with the model (\ref{packetloss})\textit{
}is studied by taking \textit{Case B} as an example, where the value
$\boldsymbol{\hat{\xi}}{_{\text{o}}}\ $is from (\ref{packetloss})
taking as the ground truth and $\boldsymbol{\bar{\xi}}_{\text{o}}\left(t\right)$
from (\ref{differentialequation}). The two models with the same communication
parameters have the same input $\boldsymbol{\xi}_{\text{o}}.$ Let
us study the noise $\boldsymbol{\varepsilon}_{\text{o}}=\boldsymbol{\hat{\xi}}{_{\text{o}}}-\boldsymbol{\bar{\xi}}_{\text{o}}.$
As shown in Figure \ref{prodenfunction}, for the UAV, the noise $\boldsymbol{\varepsilon}_{\text{o}}$
is bounded, moreover, obeying the normal distribution by Kolmogorov-Smirnov
test. Therefore, \textit{Assumption 2} is reasonable. 
\begin{figure}[ptb]
\begin{centering}
\includegraphics[scale=0.6]{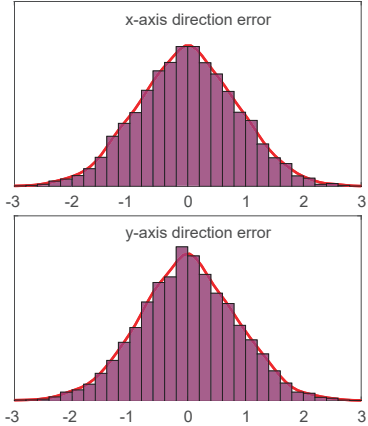} 
\par\end{centering}
\caption{Statistical property of obstacle estimate noise}
\label{prodenfunction}
\end{figure}
\item \textbf{Safety Radius Verification}.\textbf{ }Under the initial conditions
above and an obstacle avoidance controller, the \emph{true} \emph{position
distance }$\left\Vert \mathbf{\tilde{p}}_{\text{o}}\left(t\right)\right\Vert $
and the \emph{estimated filtered position distance} $\left\Vert \mathbf{e}_{\text{o}}\left(t\right)\right\Vert $
between the UAV and the obstacle are shown in Figure \ref{mindistance}.
Figure \ref{mindistance}(a) corresponding to \textit{Case A} shows
that \textit{Assumption 3} is satisfied. The results observed from
Figure \ref{flight} and Figure \ref{mindistance}(b) corresponding
to \textit{Case B}, are consistent with conclusion (iii) of \textit{Theorem
1 }and the result in\textit{ Theorem 2. }As shown in Figure \ref{flight},
it should be noted that $\left\Vert \mathbf{e}_{\text{o}}\left(t\right)\right\Vert \geq r_{\text{s}}+r_{\text{o}}$
holds consistent with conclusion (iii) in \textit{Theorem 1} during
the flight, although $\left\Vert \boldsymbol{\tilde{\xi}}{_{\text{o}}}\left(t\right)\right\Vert <r_{\text{s}}+r_{\text{o}}\ $after
$t=5$s because of uncertainties. Since \textit{Case C} does not satisfy
the condition of (\ref{noncooperconditionTh1}), as shown in Figure
\ref{mindistance}(c), $\left\Vert \mathbf{e}_{\text{o}}\left(t\right)\right\Vert <r_{\text{s}}+r_{\text{o}}$
at time about 11.25s. 
\begin{figure}[h]
\begin{centering}
\includegraphics[scale=0.6]{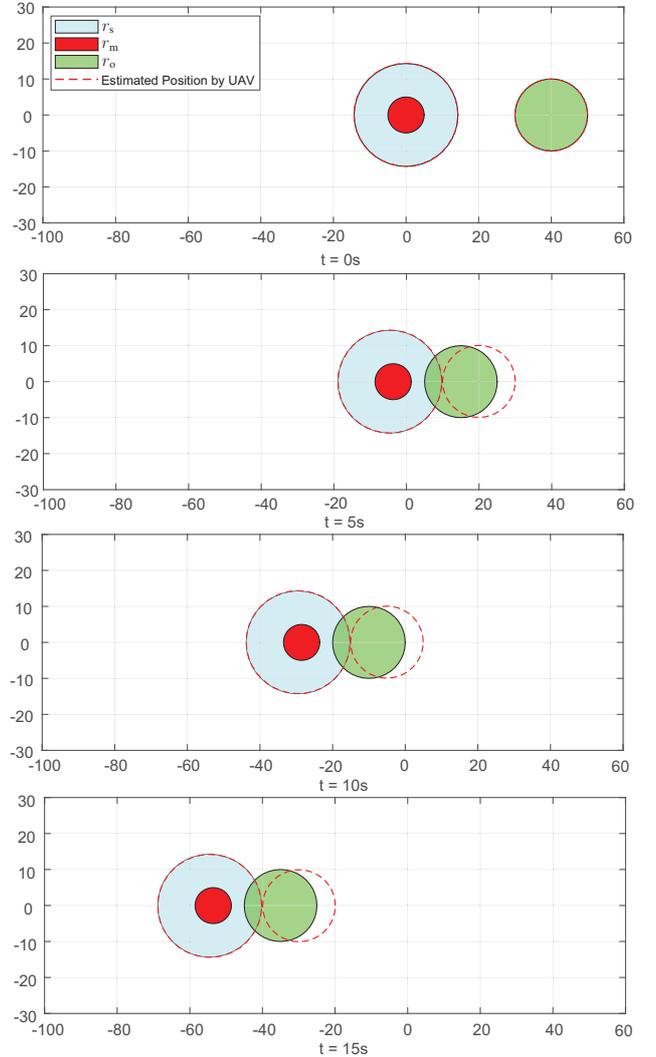} 
\par\end{centering}
\caption{Positions of UAV and obstacle at different times in Case B of numerical
simulation}
\label{flight}
\end{figure}

\begin{figure}[tp]
\begin{centering}
\includegraphics[scale=0.42]{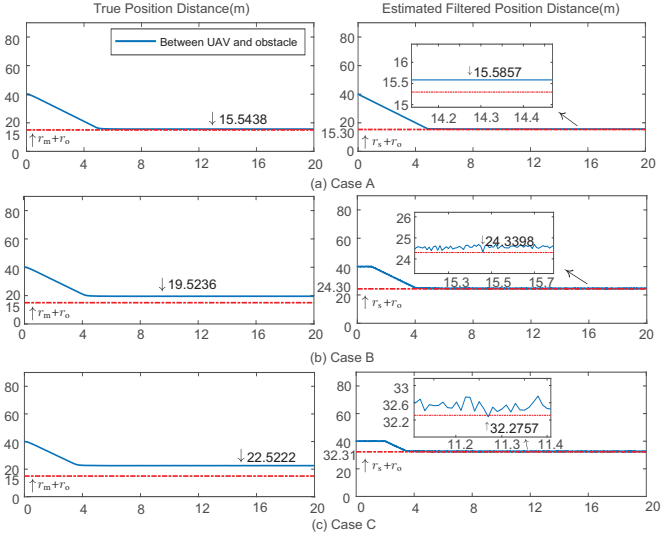} 
\par\end{centering}
\caption{The position distance and estimated filtered position distance between
a UAV and one obstacle in numerical simulation}
\label{mindistance}
\end{figure}

\end{itemize}

\subsubsection{Simulation with Multiple Non-Cooperative Obstacles}
\begin{itemize}
\item \textbf{Simulation Setting}. As shown in Figure \ref{flight-1}, a
scenario that one UAV makes avoidance with three moving non-cooperative
obstacles is considered. These obstacles can avoid each other expect
for the UAV. The simulation parameters are set as follows. The initial
position of the UAV is set as $\mathbf{p}\left(0\right)=\left[0~40~100\right]^{\text{T}}$m
with radius $r_{\text{m}}$ $=5$m; the initial positions of obstacles
are set as $\mathbf{p}_{\text{o,1}}\left(0\right)=\left[-40~-40~100\right]^{\text{T}}$m,
$\mathbf{p}_{\text{o,2}}\left(0\right)=\left[0~-40~100\right]^{\text{T}}$m,
$\mathbf{p}_{\text{o,3}}\left(0\right)=\left[40~-40~100\right]^{\text{T}}$m
with radius $r_{\text{o}}=10$m and the velocity $v_{\text{o},i}=i+2$m/s,
$i=1,2,3$. The others about the UAV and uncertainties are the same
as those in the last simulation.
\item \textbf{Safety Radius Verification}.\textbf{ }Under the initial conditions
above and an obstacle avoidance controller, the minimum\emph{ true}
\emph{position distance }$\underset{i\in\left\{ 1,2,3\right\} }{\min}\left\Vert \mathbf{\tilde{p}}_{\text{o,}i}\left(t\right)\right\Vert $
and the \emph{estimated filtered position distance} $\underset{i\in\left\{ 1,2,3\right\} }{\min}\left\Vert \mathbf{e}_{\text{o},i}\left(t\right)\right\Vert $
between the UAV and the obstacle are shown in Figure \ref{mindistance-1}.
Figure \ref{mindistance-1}(a) corresponding to \textit{Case A} shows
that \textit{Assumption 3'} is satisfied. The results observed from
Figure \ref{flight-1} and Figure \ref{mindistance-1}(b) corresponding
to \textit{Case B}, are consistent with conclusion (ii) of \textit{Theorem
4. }As shown in Figure \ref{flight-1}, it should be noted that $\underset{i\in\left\{ 1,2,3\right\} }{\min}\left\Vert \mathbf{e}_{\text{o}}\left(t\right)\right\Vert \geq r_{\text{s}}+r_{\text{o}}$
holds consistent with conclusion (ii) in \textit{Theorem 4} during
the flight, although $\underset{i\in\left\{ 1,2,3\right\} }{\min}\left\Vert \boldsymbol{\tilde{\xi}}_{\text{o,}i}\left(t\right)\right\Vert <r_{\text{s}}+r_{\text{o}}\ $about
$t=10$s because of uncertainties. Since \textit{Case C} does not
satisfy the condition of (\ref{noncooperconditionTh1}), as shown
in Figure \ref{mindistance-1}(c), $\underset{i\in\left\{ 1,2,3\right\} }{\min}\left\Vert \mathbf{e}_{\text{o},i}\left(t\right)\right\Vert <r_{\text{s}}+r_{\text{o}}$
at time about 6s. 
\begin{figure}[h]
\begin{centering}
\includegraphics[scale=0.55]{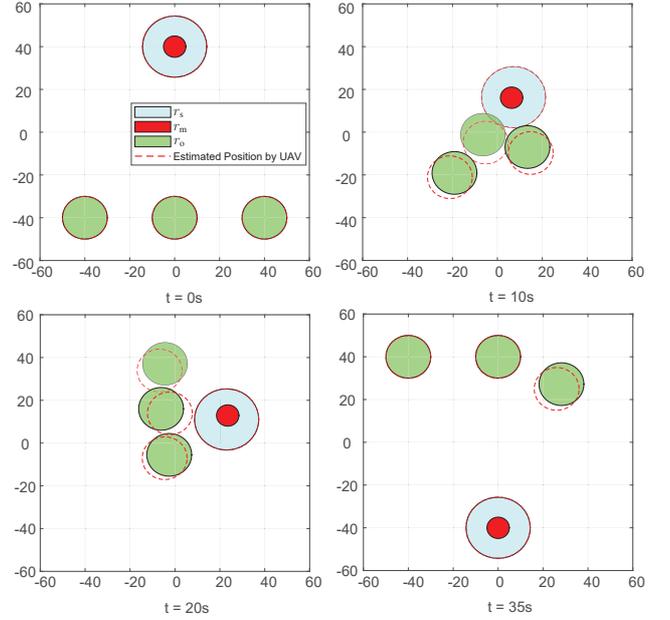} 
\par\end{centering}
\caption{Positions of UAV and three obstacles at different time in Case B of
numerical simulation}
\label{flight-1}
\end{figure}

\begin{figure}[tp]
\begin{centering}
\includegraphics[scale=0.45]{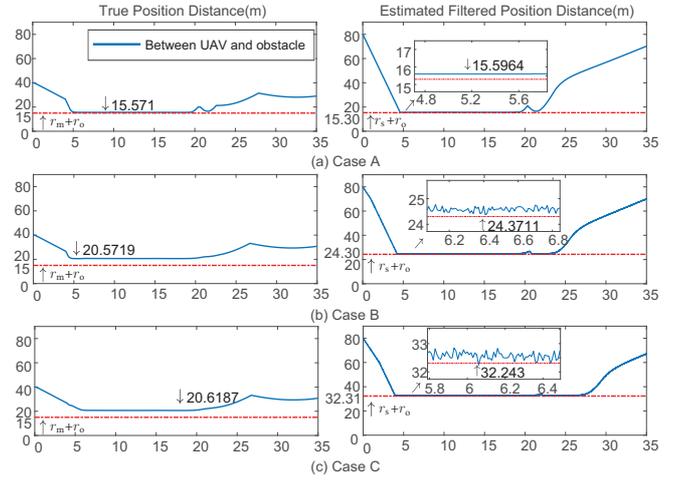} 
\par\end{centering}
\caption{Minimum position distance and estimated filtered position distance
from UAV to three non-cooperative obstacles in numerical simulation}
\label{mindistance-1}
\end{figure}

\end{itemize}

\subsubsection{Simulation with Multiple Cooperative Obstacles}
\begin{itemize}
\item \textbf{Simulation Setting}. As shown in Figure \ref{flight-1-1},
a scenario that one UAV makes avoidance with three moving cooperative
obstacles is considered. The UAV and these obstacles can avoid each
other. The simulation parameters are set as follows. The initial position
of the UAV is set as $\mathbf{p}\left(0\right)=\left[-40~40~100\right]^{\text{T}}$m
with radius $r_{\text{m}}$ $=5$m. The UAV's maneuver constant is
$l=5$, and the maximum speed $v_{\text{m}}=5\text{m/s}$. The interval
of receiving information for the UAV is $T_{\text{s}}=0.01$s. The
initial positions of obstacles are set as $\mathbf{p}_{\text{o,1}}\left(0\right)=\left[40~40~100\right]^{\text{T}}$m,
$\mathbf{p}_{\text{o,2}}\left(0\right)=\left[40~-40~100\right]^{\text{T}}$m,
$\mathbf{p}_{\text{o,3}}\left(0\right)=\left[-40~-40~100\right]^{\text{T}}$m
with radius $r_{\text{o}}=10$m and their velocities $v_{\text{o},i}=i+2$m/s,
$i=1,2,3$. Communication uncertainty parameters are set as \textit{Case
B }in Table I. The designed safety radius is chosen as $r_{\text{s}}=14.14$m
according to (\ref{Th11}) in \textit{Theorem 2. }It is worth noting
that the condition of (\ref{noncooperconditionTh1})\ does not satisfy
because of $v_{\text{m}}=v_{\text{o},3}$.
\item \textbf{Safety Radius Verification}. Under the initial conditions
above and an obstacle avoidance controller, the minimum \emph{true}
\emph{position distance }$\underset{i\in\left\{ 1,2,3\right\} }{\min}\left\Vert \mathbf{\tilde{p}}_{\text{o,}i}\left(t\right)\right\Vert $
and the \emph{estimated filtered position distance} $\underset{i\in\left\{ 1,2,3\right\} }{\min}\left\Vert \mathbf{e}_{\text{o},i}\left(t\right)\right\Vert $
from the UAV to these obstacles are shown in Figure \ref{mindistance-1-1}.
Since the obstacles and the UAV can make collision avoidance with
each other, the separation principle still holds even if $v_{\text{m}}=v_{\text{o},3}$.
For a simple case, \emph{Remark 5} has explained the reason. Consequently,
$\underset{i\in\left\{ 1,2,3\right\} }{\min}\left\Vert \mathbf{e}_{\text{o}}\left(t\right)\right\Vert \geq r_{\text{s}}+r_{\text{o}}$
holds during the flight observed from Figure \ref{flight-1-1}. 
\begin{figure}[h]
\begin{centering}
\includegraphics[scale=0.6]{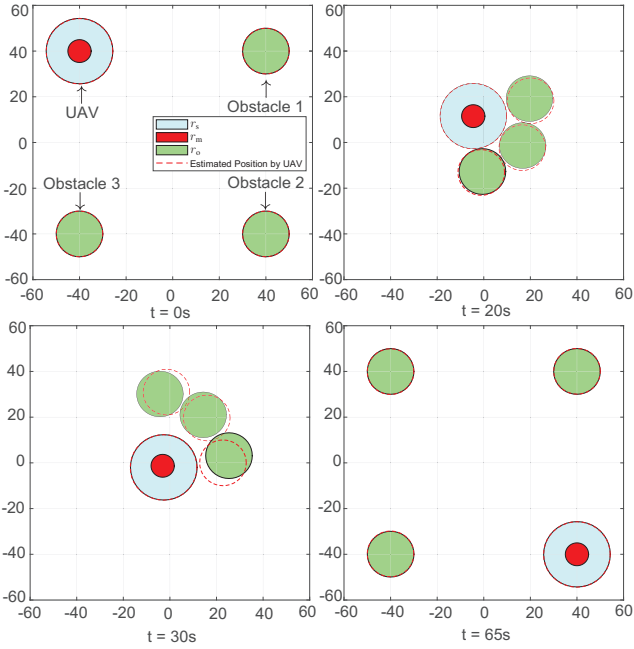} 
\par\end{centering}
\caption{Positions of UAV and three obstacles at different times in Case B
of numerical simulation}
\label{flight-1-1}
\end{figure}

\begin{figure}[ptb]
\begin{centering}
\includegraphics[scale=0.48]{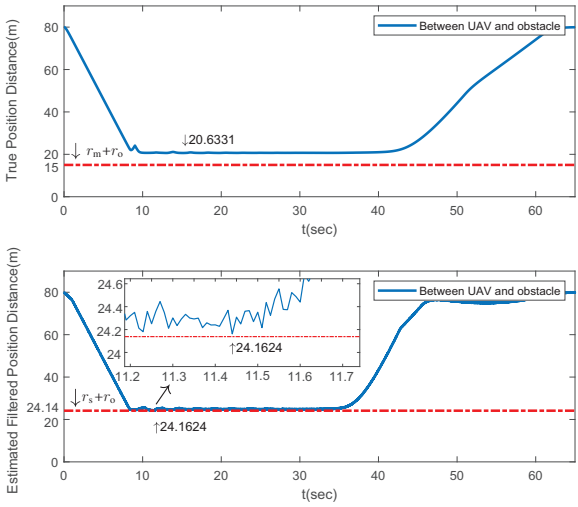} 
\par\end{centering}
\caption{Minimum position distance and filtered position distance from UAV
to three cooperative obstacles in numerical simulation}
\label{mindistance-1-1}
\end{figure}

\end{itemize}

\subsection{Experiments}

A motion capture system called OptiTrack is installed, from which
we can get the ground truth of the position, velocity and, orientation
of each multicopter. The laptop is connected to these multicopters
and OptiTrack by a local network, providing the proposed controller
and a real-time position plotting module. In the first experiment
for a non-cooperative obstacle,\ with the proposed separation principle
(\textit{Theorem 1}) and designed safety radius (\textit{Theorem 2}),
we will show that the condition (\ref{noncooperconditionTh1}) can
make avoidance subject to uncertainties. However, for cooperative
obstacles which can make avoidance simultaneously, the condition (\ref{noncooperconditionTh1})
is not necessary, which is shown in the last two experiments. The
results of \textit{Theorems 3,5 }can be observed directly from these
results of the following experiments by choosing $\hat{r}_{\text{s}}=r_{\text{s}}.$

\subsubsection{Experiment with One Non-Cooperative Hovering Obstacle}
\begin{itemize}
\item \textbf{Experiment Setting}. An experiment scenario that one UAV makes
avoidance with one non-cooperative hovering obstacle is considered.
The experiment parameters are set as follows. The UAV with physical
radius $r_{\text{m}}$ $=0.2$m is at $\mathbf{p}\left(0\right)=\left[1.5~0\text{ }1\right]^{\text{T}}$m
initially. The UAV's maneuver constant is $l=2$, and the maximum
speed $v_{\text{m}}=0.1\text{m/s}$. The obstacle is at $\mathbf{p}_{\text{o}}\left(0\right)=\left[-0.2~0\text{ }1\right]^{\text{T}}$m
initially with radius $r_{\text{o}}=0.2$m. The interval of receiving
information for the UAV is $T_{\text{s}}=0.01$s. Communication uncertainty
parameters are set as $b=0.10$m, $b_{\text{o}}=0.03$m, $v_{b}=0.08$m/s,
$v_{b_{\text{o}}}=0.01$m/s, $\tau_{\text{d}}=1$s, $\theta=10\%$.
The condition of (\ref{noncooperconditionTh1}) in \textit{Theorem
2 }is satisfied in this scenario. The safety radius $r_{\text{s}}=0.47\text{m}$
is designed with such defined parameters according to \textit{Theorem
2}.
\item \textbf{Safety Radius Verification}.\textbf{ }Under the initial conditions
above and an obstacle avoidance controller, the true position distance
and the estimated filtered position distance from the UAV to the obstacle
are shown in Figure \ref{mindis_exp1}. The positions of multicopters
during the whole flight experiment are shown in Figure \ref{position_env-1}.
The UAV can complete its route at about 77s, keeping a safe distance
from the obstacle without conflict. This is consistent with the separation
principle (\textit{Theorem 1}) with the designed safety radius (\textit{Theorem
2}). 
\begin{figure}[ptb]
\begin{centering}
\includegraphics[scale=0.5]{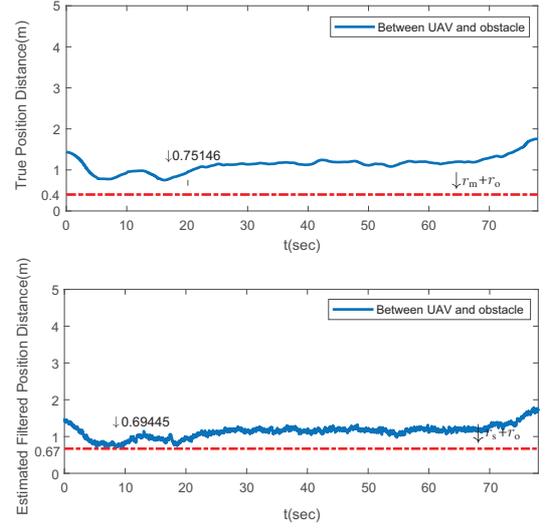} 
\par\end{centering}
\caption{The position distance and estimated filtered position distance between
a UAV and one hovering obstacle in the flight experiment}
\label{mindis_exp1}
\end{figure}

\begin{figure}
\begin{centering}
\includegraphics[scale=0.55]{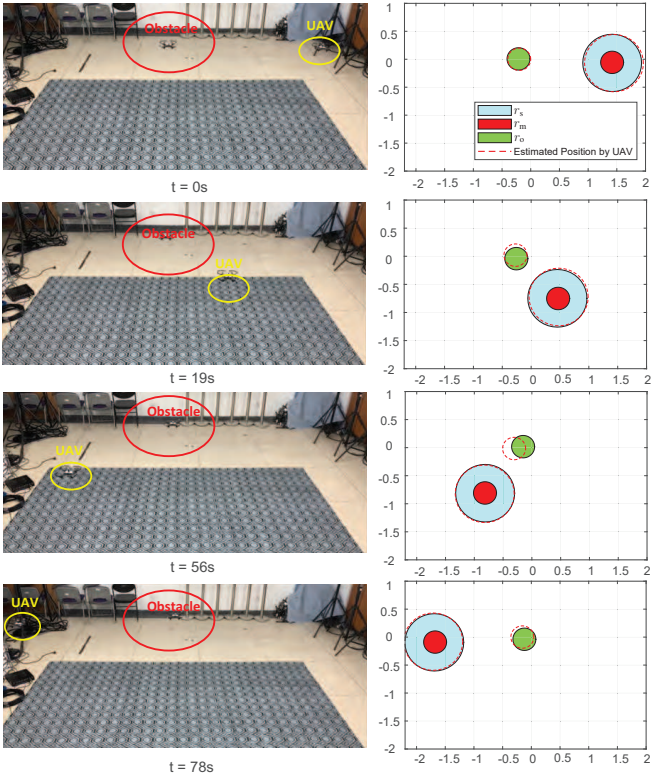} 
\par\end{centering}
\caption{Positions of a UAV and a non-cooperative hovering obstacle at different
times in the flight experiment}
\label{position_env-1}
\end{figure}

\end{itemize}

\subsubsection{Experiment with One Cooperative Moving Obstacle}
\begin{itemize}
\item \textbf{Experiment Setting}. An experiment scenario that one static
UAV makes avoidance with one moving cooperative obstacle is considered.
The experiment parameters are set as follows. The UAV with physical
radius $r_{\text{m}}$ $=0.2$m is at $\mathbf{p}\left(0\right)=\left[1.5~0\text{ }1\right]^{\text{T}}$m
initially. The UAV's maneuver constant is $l=2$, and the maximum
speed $v_{\text{m}}=0.1\text{m/s}$. The obstacle is at $\mathbf{p}_{\text{o}}\left(0\right)=\left[-1.5~0\text{ }1\right]^{\text{T}}$m
initially with radius $r_{\text{o}}=0.2$m and $v_{\text{o}}=0.1$m/s.
The interval of receiving information for the UAV is $T_{\text{s}}=0.01$s.
Communication uncertainty parameters are set as $b=0.2$m, $b_{\text{o}}=0.1$m,
$v_{b}=0.08$m/s, $v_{b_{\text{o}}}=0.01$m/s, $\tau_{\text{d}}=2$s,
$\theta=30\%$. The safety radius $r_{\text{s}}=0.71\text{m}$ is
designed with such defined parameters by \textit{Theorem 4}.
\item \textbf{Safety Radius Verification}.\textbf{ }Under the initial conditions
above and an obstacle avoidance controller, the true position distance
and the estimated filtered position distance from the UAV to the obstacle
are shown in Figure \ref{mindis_exp2}. The positions of multicopters
during the whole flight experiment are shown in Figure \ref{position_env}.
Since the obstacle and the UAV can make collision avoidance with each
other, the separation principle still holds even if $v_{\text{m}}=v_{\text{o}}$.
\emph{Remark 5} has explained the reason. This is consistent with
the separation principle (\textit{Theorem 1}) with the designed safety
radius (\textit{Theorem 2}). 
\begin{figure}[ptb]
\begin{centering}
\includegraphics[scale=0.5]{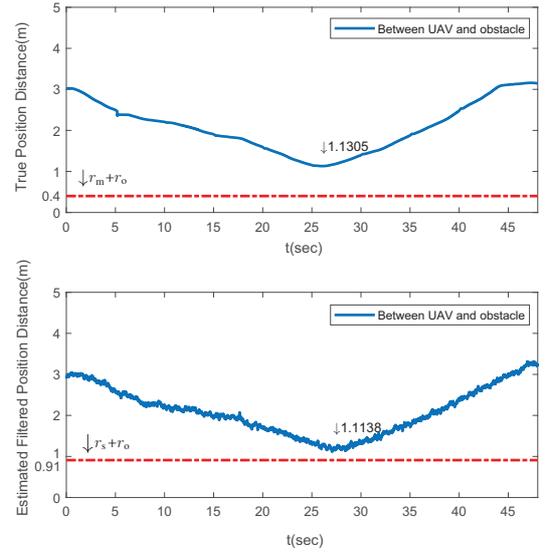} 
\par\end{centering}
\caption{The position distance and estimated filtered position distance between
a UAV and one moving obstacle in the flight experiment}
\label{mindis_exp2}
\end{figure}

\begin{figure}[ptb]
\begin{centering}
\includegraphics[scale=0.5]{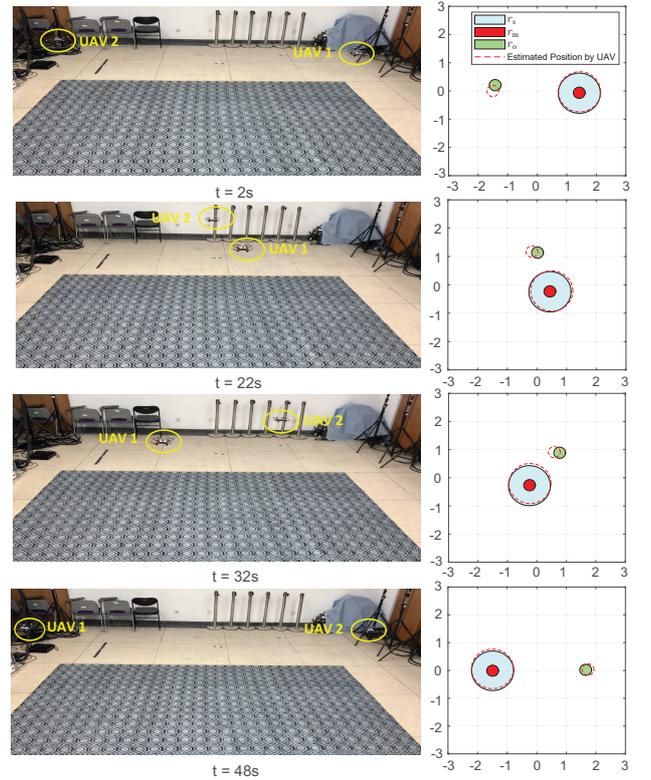} 
\par\end{centering}
\caption{Positions of UAV and a moving cooperative\ obstacle at different
times in the flight experiment}
\label{position_env}
\end{figure}

\end{itemize}

\subsubsection{Experiment with Multiple Cooperative Obstacles}
\begin{itemize}
\item \textbf{Experiment Setting}. An experiment scenario that one UAV makes
avoidance with three moving cooperative obstacles is considered. The
experiment parameters are set as follows. The UAV with a physical
radius $r_{\text{m}}$ $=0.2$m is at $\mathbf{p}\left(0\right)=\left[-1~1~1\right]^{\text{T}}$m
initially. The UAV's maneuver constant is $l=2$, and the maximum
speed $v_{\text{m}}=0.1\text{m/s}$. The obstacles are at $\mathbf{p}_{\text{o},1}\left(0\right)=\left[1~1~1\right]^{\text{T}}$m,
$\mathbf{p}_{\text{o},2}\left(0\right)=\left[1~-1~1\right]^{\text{T}}$m,
$\mathbf{p}_{\text{o},3}\left(0\right)=\left[-1~-1~1\right]^{\text{T}}$m
initially with radius $r_{\text{o}}=0.23$m and $v_{\text{o}}=0.1$m/s.
The interval of receiving information for the UAV is $T_{\text{s}}=0.01$s.
Communication uncertainty parameters are set as $b=0.012$m, $b_{\text{o}}=0.01$m,
$v_{b}=0.012$m/s, $v_{b_{\text{o}}}=0.01$m/s, $\tau_{\text{d}}=0.1$s,
$\theta=1\%$. The safety radius $r_{\text{s}}=0.23\text{m}$ is designed
according to \textit{Theorem 2}.
\item \textbf{Safety Radius Verification}. Under the initial conditions
above and an obstacle-avoidance controller, the minimum\emph{ true}
\emph{position distance }$\underset{i\in\left\{ 1,2,3\right\} }{\min}\left\Vert \mathbf{\tilde{p}}_{\text{o,}i}\left(t\right)\right\Vert $
and the \emph{estimated filtered position distance} $\underset{i\in\left\{ 1,2,3\right\} }{\min}\left\Vert \mathbf{e}_{\text{o},i}\left(t\right)\right\Vert $
from the UAV to these obstacles are shown in Figure \ref{flightpos}.
Since these obstacles and the UAV can make collision avoidance with
each other, the separation principle still holds even if $v_{\text{m}}=v_{\text{o}}$.
Consequently, $\underset{i\in\left\{ 1,2,3\right\} }{\min}\left\Vert \mathbf{e}_{\text{o}}\left(t\right)\right\Vert \geq r_{\text{s}}+r_{\text{o}}$
holds during the flight observed from Figure \ref{flightpos}.

\begin{figure}[ptb]
\begin{centering}
\includegraphics[scale=0.5]{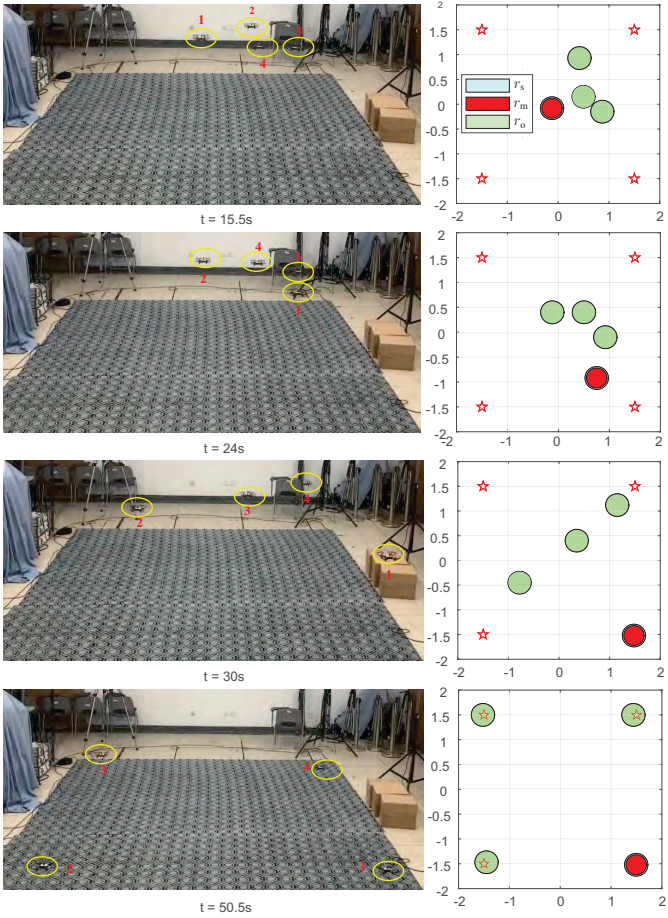}
\par\end{centering}
\caption{Positions of four UAVs at different time in flight experiment}
\label{flightpos}
\end{figure}

\end{itemize}

\subsection{Discussion}

In \emph{Assumption 3}, we proposed some requirements on the controller
performance without considering the communication uncertainties. For
multi-agent systems, the current control methods mainly include the
\emph{optimal trajectory method} and \emph{forcefield-based method}.
By simulation and analysis, the performance of different control methods
subject to communication uncertainties is shown.

\subsubsection{Forcefield-Based Control without Safety Radius Design}

\begin{figure}[tp]
\begin{centering}
\includegraphics[scale=0.4]{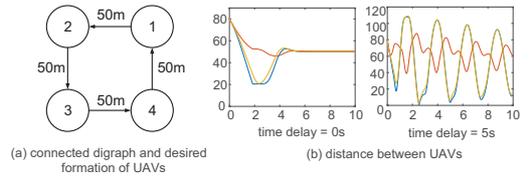}
\par\end{centering}
\caption{Performance of formation control method subject to communication uncertainties}
\label{formationcontrol}
\end{figure}

The impact of communication uncertainties on forcefield-based control
is first shown by an example of formation control. The objective of
the formation control method requires agents maintaining a certain
formation. In \cite{Oh(2015)}, existing results of the formation
control were introduced and categorized. We choose a simple and classical
displacement-based control method to show the ability of the formation
control method subject to communication uncertainties. The UAV's maneuver
constant is $l=5$, and the maximum speed $v_{\text{m}}=10\text{m/s}$.
The initial positions of UAVs are set as $\mathbf{p}_{\text{1}}\left(0\right)=\left[-40~40~100\right]^{\text{T}}$m,
$\mathbf{p}_{2}\left(0\right)=\left[40~40~100\right]^{\text{T}}$m,
$\mathbf{p}_{3}\left(0\right)=\left[40~-40~100\right]^{\text{T}}$m,
$\mathbf{p}_{4}\left(0\right)=\left[-40~-40~100\right]^{\text{T}}$m.
The communication topology and desired formation of UAVs are shown
in Figure \ref{formationcontrol}(a). For simplicity, we only consider
the impact of time delay on formation. As shown in Figure \ref{formationcontrol}(b),
the UAVs cannot converge to the desired formation in a short time
subject to time delay. The result is similar to \cite{Olfati-Saber(2004)},
which indicates that this objective is difficult to achieve if the
communication uncertainties are not compensated for elaborately in
formation controllers.

\begin{figure}[tp]
\begin{centering}
\includegraphics[scale=0.7]{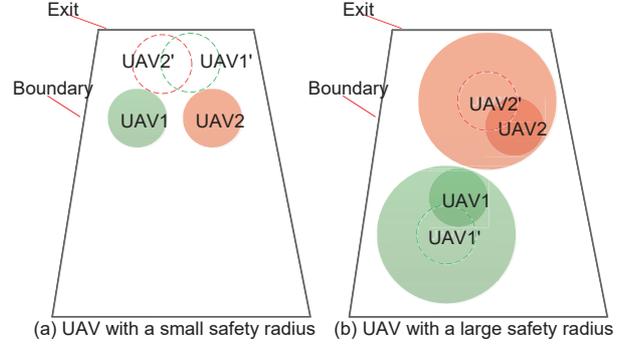}
\par\end{centering}
\caption{Positions of two UAVs without and with the designed safety radius}
\label{deadlock}
\end{figure}

Estimating-and-then-compensating is a way to deal with uncertainty.
But if the noise, delay, or packet loss is not compensated for elaborately,
phenomena like deadlock will happen. Let us consider a simple but
particular example that two UAVs pass a trapezoid tunnel. As shown
in Figure \ref{deadlock}(a), in the presence of uncertainties, UAV1
considers UAV2 at the position of UAV2\textquoteright , while UAV2
considers UAV1 at the position of UAV1\textquoteright . In this case,
a deadlock will exist, namely, each one cannot pass the exit. Even
if a deadlock does not exist, these uncertainties will slow down the
movement of the swarm. The proposed safety radius for uncertainties
can solve this problem by separating the two UAVs large enough as
shown in Figure \ref{deadlock}(b).

\subsubsection{Calculation Speed Analysis of Optimal Trajectory Method \& Forcefield-Based
Method with Safety Radius Design}

\begin{figure}[ptbh]
\begin{centering}
\includegraphics[scale=0.7]{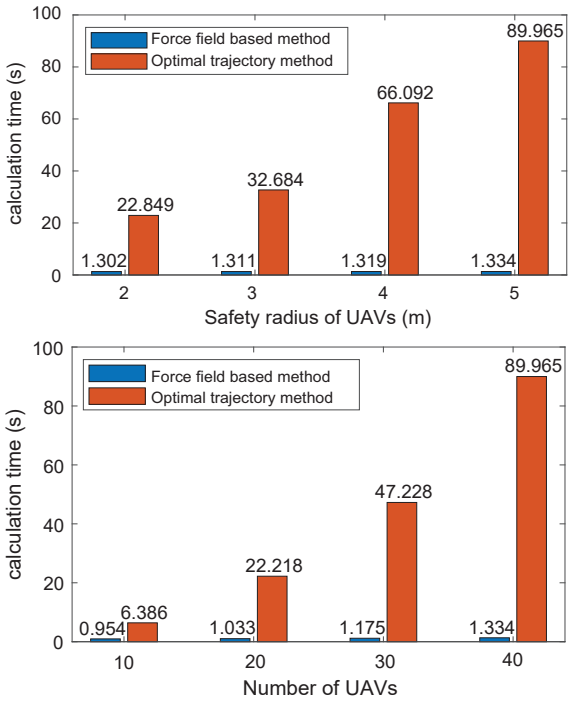}
\par\end{centering}
\caption{Calculation speed of different algorithms}
\label{compare}
\end{figure}

We discuss the performance of different control methods with the safety
radius design. The objective of the optimal trajectory method requires
optimal solutions of length, time, or energy of path, which leads
to longer calculation time of the online path-planning problem. In
the simulation, we compare the online path-planning calculation speed
of an optimization-based algorithm with the forcefield-based method
by MATLAB. In \cite{Ingersoll(2016)}, a path-planning algorithm using
Bezier curves with the open-source code at \url{https://github.com/byuflowlab/uav-path-optimization}
is proposed, which can find the optimal solutions to the offline and
online path-planning problem. We design a scenario that contains 10
UAVs at the same altitude with $r_{\text{s}}=5$m. The initial position
of 1st UAV $\mathbf{p}_{\text{1}}\left(0\right)=\left[0~0~100\right]^{\text{T}}$m,
while the other UAVs are distributed randomly in a $100\text{m}\times100\text{m}$
space with a constant velocity $[1\ 0\ 0]^{\text{T}}$. For two different
algorithms, we design 10 sets of random initial positions for the
other UAVs, run the simulation on the same computer, and record the
\emph{average calculation time} when the 1st UAV arrives at its destination
$\left[100~0~100\right]^{\text{T}}$m. Figure \ref{compare} shows
the calculation speed performance with respect to the density by changing
the safety radius and the number of UAVs separately. As shown in Figure
\ref{compare}, for the same airspace, if the number of UAVs increases
or the safety radius of UAVs gets larger, the calculation speed of
the optimization-based algorithm will be decreased rapidly because
the probability of constraint being triggered is increasing, which
brings more complex calculations. On the contrary, the forcefield-based
method can better deal with such an online path-planning problem.
However, the optimal trajectory method is better to deal with the
offline path-planning problem.

\section{Conclusions}

How to decide the safety radius taking communication uncertainties
into consideration is studied in this paper. First, a VTOL UAV model
and obstacle model are introduced. Then, some assumptions of communication
and control are made, including estimated noise, broadcast delay,
and packet loss. Based on models and assumptions, problems are formulated
to determine the designed safety radius in the design phase objective
and the practical safety radius in the flight phase objective. For
the first objective, the principle of separation of control and safety
radius (\textit{Theorem 1}) is proposed. With this principle, the
designed safety radius is determined in \textit{Theorems 2,4. }Then\textit{,
}the practical safety radius is determined in \textit{Theorems 3,5}.
By the proposed methods, a UAV can keep a safe distance from other
obstacles during the whole flight. This is very necessary to guarantee
flight safety in practice. Simulations and experiments are given to
show the effectiveness of the proposed method from the functional
requirement and the safety requirement.

\section{Appendix}

\subsection{Proof of Lemma 1}

(i) \textit{Proof of Sufficiency}. Since 
\[
\boldsymbol{\dot{\tilde{\xi}}}{_{\text{o}}}=\mathbf{c}\left(t,\boldsymbol{\tilde{\xi}}{}_{\text{o}}\right)-\boldsymbol{\dot{\xi}}_{\text{o}}
\]
we have 
\begin{equation}
\boldsymbol{\tilde{\xi}}_{\text{o}}^{\text{T}}\boldsymbol{\dot{\tilde{\xi}}}{}_{\text{o}}=\boldsymbol{\tilde{\xi}}_{\text{o}}^{\text{T}}\left(\mathbf{c}\left(t,\boldsymbol{\tilde{\xi}}{}_{\text{o}}\right)-\boldsymbol{\dot{\xi}}_{\text{o}}\right).\label{normderivlemma1}
\end{equation}
If (\ref{Safetyaera}) is violated, due to the continuity of $\boldsymbol{\tilde{\xi}}_{\text{o}}$,
there must exist a time $t=t_{1}$ such that $\left\Vert \boldsymbol{\tilde{\xi}}_{\text{o}}^{\text{T}}\left(t_{1}\right)\boldsymbol{\tilde{\xi}}_{\text{o}}\left(t_{1}\right)\right\Vert =\left(r_{\text{s}}+r_{\text{o}}\right)^{2}$.
Since $\left\Vert \boldsymbol{\dot{\xi}}{}_{\text{o}}\right\Vert \leq v_{\text{m}},$
(\ref{normderivlemma1}) becomes 
\begin{align*}
\left.\boldsymbol{\tilde{\xi}}_{\text{o}}^{\text{T}}\boldsymbol{\dot{\tilde{\xi}}}{}_{\text{o}}\right\vert _{t=t_{1}} & \geq\boldsymbol{\tilde{\xi}}_{\text{o}}^{\text{T}}\mathbf{c}\left(t_{1},\boldsymbol{\tilde{\xi}}{}_{\text{o}}\right)-\left\Vert \boldsymbol{\tilde{\xi}}_{\text{o}}\left(t_{1}\right)\right\Vert \left\Vert \boldsymbol{\dot{\xi}}{}_{\text{o}}\left(t_{1}\right)\right\Vert \\
 & \geq\boldsymbol{\tilde{\xi}}_{\text{o}}^{\text{T}}\mathbf{c}\left(t_{1},\boldsymbol{\tilde{\xi}}{}_{\text{o}}\right)-\left(r_{\text{s}}+r_{\text{o}}\right)v_{\text{m}}.
\end{align*}
If (\ref{suffnece}) holds, then $\left.\boldsymbol{\tilde{\xi}}_{\text{o}}^{\text{T}}\boldsymbol{\dot{\tilde{\xi}}}_{\text{o}}\right\vert _{t=t_{1}}\geq0.$
This implies that $\left\Vert \boldsymbol{\tilde{\xi}}_{\text{o}}^{\text{T}}\left(t_{1}\right)\boldsymbol{\tilde{\xi}}_{\text{o}}\left(t_{1}\right)\right\Vert $
will not be decreased any more. So, (\ref{Safetyaera}) cannot be
violated for any $\left\Vert \boldsymbol{\tilde{\xi}}{_{\text{o}}}\left(0\right)\right\Vert \geq r_{\text{s}}+r_{\text{o}}$.
(ii) \textit{Proof of Necessity}. This necessary condition is proved
by contradiction. Suppose that there exists an $\mathbf{x}^{\mathbf{\ast}}\in\mathcal{C}$
such that 
\begin{align*}
\mathbf{x}^{\ast\text{T}}\mathbf{c}\left(t,\mathbf{x}^{\mathbf{\ast}}\right) & =\left(r_{\text{s}}+r_{\text{o}}\right)v_{\text{m}}-\epsilon\\
 & <\left(r_{\text{s}}+r_{\text{o}}\right)v_{\text{m}}
\end{align*}
where ${\epsilon>0.}$ We will show that (\ref{Safetyaera}) will
not hold for any obstacle $\left\Vert \boldsymbol{\dot{\xi}}_{\text{o}}\right\Vert \leq{v_{\text{m}}\ }${and}$\ \left\Vert \boldsymbol{\tilde{\xi}}{_{\text{o}}}\left(0\right)\right\Vert \geq r_{\text{s}}+r_{\text{o}}.$
Let $\boldsymbol{\tilde{\xi}}_{\text{o}}\left(0\right)=\mathbf{x}^{\mathbf{\ast}}\ $and
\[
\boldsymbol{\dot{\xi}}{_{\text{o}}}\left(0\right)=\frac{{v_{\text{m}}}}{r_{\text{s}}+r_{\text{o}}}\boldsymbol{\tilde{\xi}}_{\text{o}}\left(0\right).
\]
So, $\left\Vert \boldsymbol{\dot{\xi}}{_{\text{o}}}\left(0\right)\right\Vert =\frac{{v_{\text{m}}}}{r_{\text{s}}+r_{\text{o}}}\left\Vert \boldsymbol{\tilde{\xi}}_{\text{o}}\left(0\right)\right\Vert ={v_{\text{m}}.}$
In this case, (\ref{normderivlemma1}) becomes 
\begin{align*}
\left.\boldsymbol{\tilde{\xi}}_{\text{o}}^{\text{T}}\boldsymbol{\dot{\tilde{\xi}}}{_{\text{o}}}\right\vert _{t=0} & =\left(r_{\text{s}}+r_{\text{o}}\right){v_{\text{m}}}-{\epsilon}-\frac{{v_{\text{m}}}}{r_{\text{s}}+r_{\text{o}}}\boldsymbol{\tilde{\xi}}_{\text{o}}^{\text{{T}}}\left(0\right)\boldsymbol{\tilde{\xi}}_{\text{o}}\left(0\right)\\
 & =-\epsilon<0.
\end{align*}
This implies that $\left\Vert \boldsymbol{\tilde{\xi}}_{\text{o}}\left(t\right)\right\Vert $
will be further decreased around $t=0$, namely there exists a $t=t_{2}$
such that $\left\Vert \boldsymbol{\tilde{\xi}}{_{\text{o}}}\left(t_{2}\right)\right\Vert <r_{\text{s}}+r_{\text{o}}$.
This contradicts with (\ref{Safetyaera}). So, (\ref{suffnece}) is
also necessary. $\square$

\subsection{Proof of Lemma 2}

First, multiplying $\mathbf{x}^{\text{T}}\left(t\right)$ on the left
side of (\ref{equlemma1}) results in 
\begin{equation}
\mathbf{x}^{\text{T}}\left(t\right)\mathbf{\dot{x}}\left(t\right)=-k\left(t\right)\mathbf{x}^{\text{T}}\left(t\right)\mathbf{x}\left(t\right)+k\left(t\right)\mathbf{x}^{\text{T}}\left(t\right)\mathbf{y}\left(t\right).\label{equlemma2}
\end{equation}
Since 
\begin{align}
\mathbf{x}^{\text{T}}\left(t\right)\mathbf{\dot{x}}\left(t\right) & =\frac{1}{2}\frac{\text{d}\mathbf{x}^{\text{T}}\left(t\right)\mathbf{x}\left(t\right)}{\text{d}t}\nonumber \\
 & =\frac{1}{2}\frac{\text{d}\left\Vert \mathbf{x}\left(t\right)\right\Vert ^{2}}{\text{d}t}=\left\Vert \mathbf{x}\left(t\right)\right\Vert \frac{\text{d}\left\Vert \mathbf{x}\left(t\right)\right\Vert }{\text{d}t}\label{equlemma21}
\end{align}
the equation (\ref{equlemma2}) becomes 
\[
\frac{\text{d}\left\Vert \mathbf{x}\left(t\right)\right\Vert }{\text{d}t}=-k\left(t\right)\left\Vert \mathbf{x}\left(t\right)\right\Vert +k\left(t\right)\frac{1}{\left\Vert \mathbf{x}\left(t\right)\right\Vert }\mathbf{x}^{\text{T}}\left(t\right)\mathbf{y}\left(t\right).
\]
If $\left\Vert \mathbf{y}\left(t\right)\right\Vert \leq y_{\max},$
then 
\[
\frac{\text{d}\left\Vert \mathbf{x}\left(t\right)\right\Vert }{\text{d}t}\leq-k\left(t\right)\left\Vert \mathbf{x}\left(t\right)\right\Vert +k\left(t\right)y_{\max}.
\]
Let $z\triangleq\left\Vert \mathbf{x}\right\Vert -y_{\max}.$ Then
\[
\frac{\text{d}z\left(t\right)}{\text{d}t}\leq-k\left(t\right)z\left(t\right).
\]
We consider another equation that 
\[
\frac{\text{d}z^{\prime}\left(t\right)}{\text{d}t}=-k\left(t\right)z^{\prime}\left(t\right),z^{\prime}\left(0\right)=z\left(0\right).
\]
The solution to the equation above is 
\begin{equation}
z^{\prime}\left(t\right)=e^{%TCIMACRO{\dint\nolimits _{0}^{t}}%
%BeginExpansion
{\displaystyle \int\nolimits _{0}^{t}}%EndExpansion
-k\left(s\right)\text{d}s}z\left(0\right).\label{z'}
\end{equation}
Since $z\left(0\right)\leq0,$ we have $z^{\prime}\left(t\right)\leq0$
according to (\ref{z'}). Consequently, $z\left(t\right)\leq z^{\prime}\left(t\right)\leq0$
according to the comparison lemma \cite{Khalil(2002)}, namely $\left\Vert \mathbf{x}\left(t\right)\right\Vert \leq y_{\max}$.

In the following, the conclusion (\ref{conlemma12}) will be shown.
Let $\mathbf{z}\triangleq\mathbf{x}-\mathbf{y}$. Then (\ref{equlemma1})
can be transformed as 
\[
\mathbf{\dot{z}}\left(t\right)=-k\left(t\right)\mathbf{z}\left(t\right)+k\left(t\right)\left(\frac{1}{k\left(t\right)}\mathbf{\dot{y}}\right).
\]
If $\left\Vert \mathbf{z}\left(0\right)\right\Vert \leq\frac{1}{k_{\min}}v_{y_{\max}},$
then 
\begin{equation}
\left\Vert \mathbf{z}\left(t\right)\right\Vert \leq\frac{1}{k_{\min}}v_{y_{\max}}\label{lemma1z}
\end{equation}
where the conclusion (\ref{conlemma11}) is utilized. The equation
(\ref{equlemma1}) is further written as 
\[
\mathbf{\dot{x}}\left(t\right)=-k\left(t\right)\mathbf{z}\left(t\right).
\]
It can be further written as 
\begin{align*}
\left\Vert \mathbf{\dot{x}}\left(t\right)\right\Vert  & \leq\left\vert k\left(t\right)\right\vert \left\Vert \mathbf{z}\left(t\right)\right\Vert \\
 & \leq k_{\max}\left\Vert \mathbf{z}\left(t\right)\right\Vert .
\end{align*}
Using (\ref{lemma1z}) will lead to the conclusion (\ref{conlemma12}).
$\square$

\subsection{Proof of Proposition 2}

\textit{(i) Proof of sufficiency.} Let 
\begin{align*}
p & =\mathbf{\tilde{p}}_{\text{o}}^{\text{T}}\mathbf{\tilde{p}}{_{\text{o}}}\\
\delta & =\boldsymbol{\tilde{\xi}}{_{\text{o}}^{\text{T}}}\boldsymbol{\tilde{\xi}}{_{\text{o}}}-\frac{1}{l^{2}}\mathbf{\tilde{v}}{_{\text{o}}^{\text{T}}\mathbf{\tilde{v}}{_{\text{o}}}.}
\end{align*}
According to (\ref{errors}), we have 
\begin{align}
\boldsymbol{\tilde{\xi}}{_{\text{o}}^{\text{T}}}\boldsymbol{\tilde{\xi}}{_{\text{o}}} & =\left(\mathbf{\tilde{p}}{_{\text{o}}}+\frac{1}{l}\mathbf{\tilde{v}}{_{\text{o}}}\right)^{\text{T}}\left(\mathbf{\tilde{p}}{_{\text{o}}}+\frac{1}{l}\mathbf{\tilde{v}}{_{\text{o}}}\right)\nonumber \\
 & =\mathbf{\tilde{p}}_{\text{o}}^{\text{T}}\mathbf{\tilde{p}}{_{\text{o}}}+\frac{1}{l^{2}}\mathbf{\tilde{v}}{_{\text{o}}^{\text{T}}\mathbf{\tilde{v}}{_{\text{o}}}}+\frac{2}{l}\mathbf{\tilde{v}}{_{\text{o}}^{\text{T}}\mathbf{\tilde{p}}{_{\text{o}}}.}\label{filtererror2}
\end{align}
Since 
\[
\dot{p}=2\mathbf{\tilde{p}}_{\text{o}}^{\text{T}}{\mathbf{\tilde{v}}{_{\text{o}}}}
\]
using the equation (\ref{filtererror2}), we further have 
\begin{equation}
\dot{p}=-lp+l\delta.\label{equ}
\end{equation}
The solution $p\left(t\right)$ can be expressed as 
\begin{equation}
p\left(t\right)=e^{-lt}p\left(0\right)+%TCIMACRO{\dint\nolimits _{0}^{t}}%
%BeginExpansion
{\displaystyle \int\nolimits _{0}^{t}}%EndExpansion
e^{-l\left(t-s\right)}l\delta\left(s\right)\text{d}s.\label{z}
\end{equation}
With (\ref{rv1}) in hand, if condition (\ref{p3condition}) is satisfied,
then 
\[
\delta\left(t\right)=\boldsymbol{\tilde{\xi}}_{\text{o}}^{\text{T}}\boldsymbol{\tilde{\xi}}_{\text{o}}-\frac{1}{l^{2}}\mathbf{\tilde{v}}{_{\text{o}}^{\text{T}}\mathbf{\tilde{v}}{_{\text{o}}}}\geq{r^{2}.}
\]
Since $\left\Vert \mathbf{\tilde{p}}{_{\text{o}}}\left(0\right)\right\Vert >r,$
we have $p\left(0\right)>r^{2}.$ The solution in (\ref{z}) satisfies
\begin{align*}
p\left(t\right) & \geq e^{-lt}r^{2}+{r^{2}}%TCIMACRO{\dint\nolimits _{0}^{t}}%
%BeginExpansion
{\displaystyle \int\nolimits _{0}^{t}}%EndExpansion
e^{-l\left(t-s\right)}l\text{d}s\\
 & =r^{2}.
\end{align*}
Based on it, we have $\left\Vert \mathbf{\tilde{p}}{_{\text{o}}\left(t\right)}\right\Vert \geq r,$
where $t\geq0$. If $\frac{\mathbf{v}^{\text{T}}{{\mathbf{v}}_{\text{o}}}}{\left\Vert \mathbf{v}\right\Vert \left\Vert {{\mathbf{v}}_{\text{o}}}\right\Vert }=-1,$
then the UAV and the obstacle are in the case shown in Figure \ref{Intuitive}(b).
Thus, 
\[
\frac{1}{l^{2}}\mathbf{\tilde{v}}{_{\text{o}}^{\text{T}}\mathbf{\tilde{v}}{_{\text{o}}}}=r_{\text{v}}.
\]
Consequently, $\delta\left(t\right)={r^{2}.}$ Furthermore, if $\left\Vert \boldsymbol{\tilde{\xi}}{_{\text{o}}}\left(t\right)\right\Vert >\sqrt{r^{2}+r_{\text{v}}^{2}}$
and $\left\Vert \mathbf{\tilde{p}}{_{\text{o}}}\left(0\right)\right\Vert >r,$
then $\left\Vert \mathbf{\tilde{p}}{_{\text{o}}}\left(t\right)\right\Vert >r,$
where $t>0$.

\textit{(ii) Proof of necessity.} Given any $\epsilon_{\text{o}}>0,$
we will show if 
\[
\left\Vert \boldsymbol{\tilde{\xi}}{_{\text{o}}}\left(t\right)\right\Vert ^{2}=r^{2}+r_{\text{v}}^{2}-\epsilon_{\text{o}},
\]
and $\left\Vert \mathbf{\tilde{p}}{_{\text{o}}}\left(0\right)\right\Vert =r,$
then there exists a case that $\left\Vert \mathbf{\tilde{p}}{_{\text{o}}}\left(t\right)\right\Vert <r,$
where $t\geq0$. Consider a case $\frac{\mathbf{v}^{\text{T}}{{\mathbf{v}}_{\text{o}}}}{\left\Vert \mathbf{v}\right\Vert \left\Vert {{\mathbf{v}}_{\text{o}}}\right\Vert }=-1.$
Then the UAV and the obstacle are in the case shown in Figure \ref{Intuitive}(b).
Thus, 
\[
\delta\left(t\right)=\boldsymbol{\tilde{\xi}}_{\text{o}}^{\text{T}}\boldsymbol{\tilde{\xi}}_{\text{o}}-\frac{1}{l^{2}}\mathbf{\tilde{v}}{_{\text{o}}^{\text{T}}\mathbf{\tilde{v}}{_{\text{o}}}}={r^{2}-\epsilon_{\text{o}}.}
\]
According to (\ref{z}), we have 
\begin{align*}
p\left(t\right) & =r^{2}-%TCIMACRO{\dint\nolimits _{0}^{t}}%
%BeginExpansion
{\displaystyle \int\nolimits _{0}^{t}}%EndExpansion
e^{-l\left(t-s\right)}l\epsilon_{\text{o}}\text{d}s\\
 & <r^{2}.
\end{align*}
Therefore, $\left\Vert \mathbf{\tilde{p}}{_{\text{o}}}\left(t\right)\right\Vert <r,$
where $t\geq0$. $\square$

\subsection{Proof of Theorem 1}

\textit{Proof of Conclusion (i)}. According to (\ref{L11}) in the
proof of \textit{Lemma 1},\textbf{ }if and only if 
\begin{equation}
\left.\mathbf{e}_{\text{o}}^{\text{T}}\mathbf{\dot{e}}_{\text{o}}\right\vert _{\left\Vert \mathbf{e}_{\text{o}}\right\Vert =r_{\text{s}}+r_{\text{o}}}\geq0\label{SPTheorem1}
\end{equation}
then (\ref{eo}) holds with $\hat{r}_{\text{s}}=r_{\text{s}}\ $for
any $\left\Vert \mathbf{e}_{\text{o}}\left(0\right)\right\Vert \geq r_{\text{s}}+r_{\text{o}}.$
The derivative of $\mathbf{e}_{\text{o}}$ is 
\[
\mathbf{\dot{e}}_{\text{o}}=\boldsymbol{\dot{\xi}}-\boldsymbol{\dot{\hat{\xi}}}{_{\text{o}}}+\boldsymbol{\dot{\varepsilon}}.
\]
Then, the inequality (\ref{SPTheorem1}) is rewritten as 
\begin{equation}
\left.\left(\mathbf{e}_{\text{o}}^{\text{T}}\boldsymbol{\dot{\xi}}-\mathbf{e}_{\text{o}}^{\text{T}}\boldsymbol{\dot{\hat{\xi}}}{_{\text{o}}}+\mathbf{e}_{\text{o}}^{\text{T}}\boldsymbol{\dot{\varepsilon}}\right)\right\vert _{\left\Vert \mathbf{e}_{\text{o}}\right\Vert =r_{\text{s}}+r_{\text{o}}}\geq0.\label{SPTheorem4}
\end{equation}
i) \textit{Proof of Sufficiency}. Since $\left\Vert \boldsymbol{\dot{\varepsilon}}\right\Vert \leq v_{b},$
we have 
\begin{align*}
 & \left.\left(\mathbf{e}_{\text{o}}^{\text{T}}\boldsymbol{\dot{\xi}}-\mathbf{e}_{\text{o}}^{\text{T}}\boldsymbol{\dot{\hat{\xi}}}{_{\text{o}}}+\mathbf{e}_{\text{o}}^{\text{T}}\boldsymbol{\dot{\varepsilon}}\right)\right\vert _{\left\Vert \mathbf{e}_{\text{o}}\right\Vert =r_{\text{s}}+r_{\text{o}}}\\
 & \geq\left.\left(\mathbf{e}_{\text{o}}^{\text{T}}\boldsymbol{\dot{\xi}}-\mathbf{e}_{\text{o}}^{\text{T}}\boldsymbol{\dot{\hat{\xi}}}{_{\text{o}}-}\left\Vert \mathbf{e}_{\text{o}}\right\Vert \left\Vert \boldsymbol{\dot{\varepsilon}}\right\Vert \right)\right\vert _{\left\Vert \mathbf{e}_{\text{o}}\right\Vert =r_{\text{s}}+r_{\text{o}}}\\
 & \geq\left.\left(\mathbf{e}_{\text{o}}^{\text{T}}\boldsymbol{\dot{\xi}}-\mathbf{e}_{\text{o}}^{\text{T}}\boldsymbol{\dot{\hat{\xi}}}{_{\text{o}}}\right)\right\vert _{\left\Vert \mathbf{e}_{\text{o}}\right\Vert =r_{\text{s}}+r_{\text{o}}}-\left(r_{\text{s}}+r_{\text{o}}\right)v_{b}.
\end{align*}
Therefore, if (\ref{ConTheorem1}) holds, then (\ref{SPTheorem4})
is satisfied. ii) \textit{Proof of Necessity}. The necessary condition
is proved by contradiction. Suppose (\ref{ConTheorem1}) is not satisfied,
namely there exists an $\mathbf{e}_{\text{o}}^{\ast}$ with $\epsilon>0$
such that 
\[
\left.\left(\mathbf{e}_{\text{o}}^{\ast}{}^{\text{T}}\boldsymbol{\dot{\xi}}-\mathbf{e}_{\text{o}}^{\ast\text{T}}\boldsymbol{\dot{\hat{\xi}}}{_{\text{o}}}\right)\right\vert _{\left\Vert \mathbf{e}_{\text{o}}^{\ast}\right\Vert =r_{\text{s}}+r_{\text{o}}}=\left(r_{\text{s}}+r_{\text{o}}\right)v_{b}-\epsilon.
\]
Then, choose $\boldsymbol{\dot{\varepsilon}}=-v_{b}\frac{\mathbf{e}_{\text{o}}^{\ast}}{r_{\text{s}}+r_{\text{o}}}$
when $\mathbf{e}_{\text{o}}=\mathbf{e}_{\text{o}}^{\ast}$, which
satisfies $\left\Vert \boldsymbol{\dot{\varepsilon}}\right\Vert \leq v_{b}.$
As a result, at $\mathbf{e}_{\text{o}}=\mathbf{e}_{\text{o}}^{\ast},$
(\ref{SPTheorem4}) becomes 
\[
\left(r_{\text{s}}+r_{\text{o}}\right)v_{b}-\epsilon-v_{b}\mathbf{e}_{\text{o}}^{\ast\text{T}}\frac{\mathbf{e}_{\text{o}}^{\ast}}{r_{\text{s}}+r_{\text{o}}}\geq0
\]
namely, 
\[
-\epsilon\geq0.
\]
This is a contradiction. This is implies that (\ref{SPTheorem4})
and then (\ref{SPTheorem1}) will be violated.

\textit{Proof of Conclusion (ii)}. Under \textit{Assumption 3}, according
to \textit{Lemma 1}, we have 
\begin{equation}
\left.\mathbf{e}_{\text{o}}^{\text{T}}\mathbf{c}\left(t,\mathbf{e}_{\text{o}}\right)\right\vert _{\left\Vert \mathbf{e}_{\text{o}}\right\Vert =r_{\text{s}}+r_{\text{o}}}\geq\left(r_{\text{s}}+r_{\text{o}}\right){v_{\text{m}}.}\label{SPTheorem2}
\end{equation}
Therefore, (\ref{ConTheorem1}) becomes (\ref{cooperconditionTh1}).

\textit{Proof of Conclusion (iii)}. In view of (\ref{differentialequation}),
according to \emph{Lemma 2}, we have 
\[
\left\Vert \boldsymbol{\dot{\bar{\xi}}}_{\text{o}}\right\Vert \leq\left\Vert \boldsymbol{\dot{\xi}}_{\text{o}}\left(t-\tau_{\text{d}}\right)\right\Vert \leq{v_{\text{o}}.}
\]
Then 
\begin{align*}
\left.\mathbf{e}_{\text{o}}^{\text{T}}\boldsymbol{\dot{\hat{\xi}}}{_{\text{o}}}\right\vert _{\left\Vert \mathbf{e}_{\text{o}}\right\Vert =r_{\text{s}}+r_{\text{o}}} & \leq\left(\left\Vert \mathbf{e}_{\text{o}}\right\Vert \left\Vert \boldsymbol{\dot{\bar{\xi}}}_{\text{o}}\right\Vert +\left\Vert \mathbf{e}_{\text{o}}\right\Vert \left\Vert \boldsymbol{\varepsilon}_{\text{o}}\right\Vert \right)_{\left\Vert \mathbf{e}_{\text{o}}\right\Vert =r_{\text{s}}+r_{\text{o}}}\\
 & \leq\left(r_{\text{s}}+r_{\text{o}}\right)\left({v_{\text{o}}}+v_{b_{\text{o}}}\right){.}
\end{align*}
If (\ref{noncooperconditionTh1}) holds, then (\ref{cooperconditionTh1})
holds. Therefore, (\ref{eo}) holds with $\hat{r}_{\text{s}}=r_{\text{s}}.$
$\square$

\subsection{Proof of Proposition 3}

Since $\mathbf{e}_{\text{o}}$ in (\ref{eo1})\ can be written as
\begin{equation}
\mathbf{e}_{\text{o}}=\boldsymbol{\tilde{\xi}}_{\text{o}}+\left(\boldsymbol{\lambda}_{\text{o}}-\boldsymbol{\varepsilon}_{\text{o}}+\boldsymbol{\varepsilon}\right)\label{e0exp}
\end{equation}
where 
\[
\boldsymbol{\lambda}_{\text{o}}\triangleq\boldsymbol{\xi}_{\text{o}}-\boldsymbol{\bar{\xi}}_{\text{o}}.
\]
Taking the norm on both sides of (\ref{e0exp}) results in 
\[
\left\Vert \mathbf{e}_{\text{o}}\right\Vert \leq\left\Vert \boldsymbol{\tilde{\xi}}_{\text{o}}\right\Vert +\left\Vert \boldsymbol{\lambda}_{\text{o}}\right\Vert +\left\Vert \boldsymbol{\varepsilon}_{\text{o}}\right\Vert +\left\Vert \boldsymbol{\varepsilon}\right\Vert .
\]
If (\ref{p3_eo}) holds, then 
\begin{align}
\left\Vert \boldsymbol{\tilde{\xi}}_{\text{o}}\left(t\right)\right\Vert  & \geq r+r_{e}-\left(\left\Vert \boldsymbol{\lambda}_{\text{o}}\right\Vert +\left\Vert \boldsymbol{\varepsilon}_{\text{o}}\right\Vert +\left\Vert \boldsymbol{\varepsilon}\right\Vert \right)\nonumber \\
 & \geq r+\frac{\theta_{\text{m}}T_{\text{s}}}{1-\theta_{\text{m}}}v_{\text{o}}+v_{\text{o}}\tau_{\text{dm}}-\left\Vert \boldsymbol{\lambda}_{\text{o}}\right\Vert \label{ksio}
\end{align}
where \textit{Assumptions 1-2} are utilized. The left work is to study
$\left\Vert \boldsymbol{\lambda}_{\text{o}}\right\Vert .$ The derivative
of $\boldsymbol{\lambda}_{\text{o}}$ is 
\begin{align}
\boldsymbol{\dot{\lambda}}_{\text{o}}\left(t\right) & =\boldsymbol{\dot{\xi}}_{\text{o}}\left(t\right)+\frac{1-\theta}{\theta T_{\text{s}}}\boldsymbol{\bar{\xi}}_{\text{o}}\left(t\right)-\frac{1-\theta}{\theta T_{\text{s}}}\boldsymbol{\xi}_{\text{o}}\left(t-\tau_{\text{d}}\right)\nonumber \\
 & =-\frac{1-\theta}{\theta T_{\text{s}}}\boldsymbol{\lambda}_{\text{o}}\left(t\right)\nonumber \\
 & \text{ \ \ }+\frac{1-\theta}{\theta T_{\text{s}}}\left(\left(\boldsymbol{\xi}_{\text{o}}\left(t\right)-\boldsymbol{\xi}_{\text{o}}\left(t-\tau_{\text{d}}\right)\right)+\frac{\theta T_{\text{s}}}{1-\theta}\boldsymbol{\dot{\xi}}_{\text{o}}\left(t\right)\right).\label{difflambda}
\end{align}
As for term $\boldsymbol{\xi}_{\text{o}}\left(t\right)-\boldsymbol{\xi}_{\text{o}}\left(t-\tau_{\text{d}}\right),$
by the mean value theorem, we have 
\[
\boldsymbol{\xi}_{\text{o}}\left(t\right)-\boldsymbol{\xi}_{\text{o}}\left(t-\tau_{\text{d}}\right)=\boldsymbol{\dot{\xi}}_{\text{o}}\left(st+\left(1-s\right)\left(t-\tau_{\text{d}}\right)\right)\tau_{\text{d}},\text{ }s\in\left[0,1\right].
\]
Then 
\[
\left\Vert \boldsymbol{\xi}_{\text{o}}\left(t-\tau_{\text{d}}\right)-\boldsymbol{\xi}_{\text{o}}\left(t\right)\right\Vert \leq v_{\text{o}}\tau_{\text{d}}
\]
where $\max\left\Vert \boldsymbol{\dot{\xi}}_{\text{o}}\right\Vert \leq v_{\text{o}}\ $is
utilized. Similarly, $\left\Vert \boldsymbol{\lambda}_{\text{o}}\left(0\right)\right\Vert \leq v_{\text{o}}\tau_{\text{d}}$
by \textit{Assumption 2}. Furthermore, based on the equation (\ref{difflambda}),
according to \textit{Lemma 2}, we have 
\begin{align}
\left\Vert \boldsymbol{\lambda}_{\text{o}}\left(t\right)\right\Vert  & \leq v_{\text{o}}\tau_{\text{d}}+\frac{\theta T_{\text{s}}}{1-\theta}{v_{\text{o}}}\nonumber \\
 & \leq v_{\text{o}}\tau_{\text{dm}}+\frac{\theta_{\text{m}}T_{\text{s}}}{1-\theta_{\text{m}}}{v_{\text{o}}.}\label{lambdao}
\end{align}
Using (\ref{ksio}) and (\ref{lambdao}) yields (\ref{ksi_eo}). $\square$

\bigskip{}

\begin{IEEEbiography}[{{{{{\includegraphics[clip,width=1in,height=1.25in]{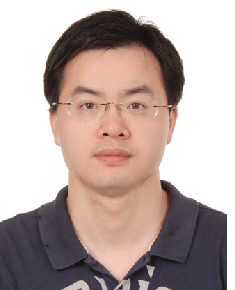}}}}}}]{Quan Quan}
received the B.S. and Ph.D. degrees in control science and engineering
from Beihang University, Beijing, China, in 2004 and 2010, respectively.
He has been an Associate Professor with Beihang University since 2013,
where he is currently with the School of Automation Science and Electrical
Engineering. His research interests include reliable flight control,
vision-based navigation, repetitive learning control, and time-delay
systems.
\end{IEEEbiography}

\begin{IEEEbiography}[{\includegraphics[clip,width=1in,height=1.25in]{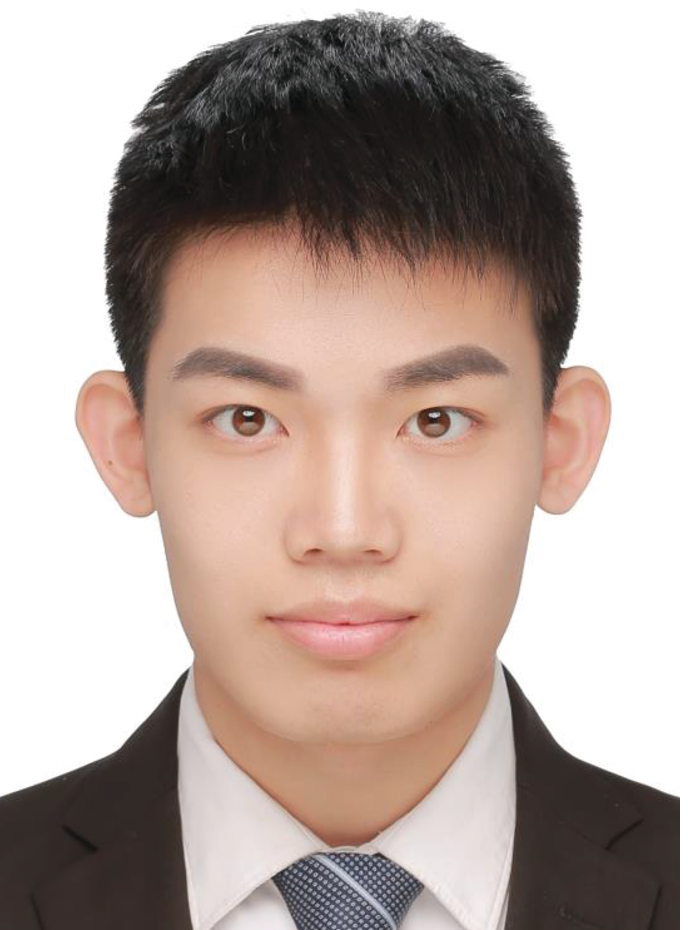}}]{Rao Fu}
received the B.S. degree in control science and engineering from
Beihang University, Beijing, China, in 2017. He is working toward
to the Ph.D. degree at the School of Automation Science and Electrical
Engineering, Beihang University (formerly Beijing University of Aeronautics
and Astronautics), Beijing, China. His main research interests include
UAV traffic control and swarm.
\end{IEEEbiography}

\begin{IEEEbiography}[{\includegraphics[clip,width=1in,height=1.25in]{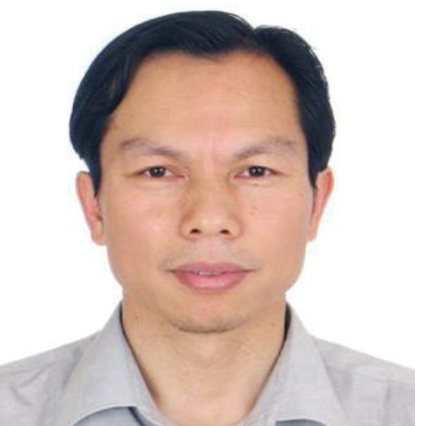}}]{Kai-Yuan Cai}
 Kai-Yuan Cai received the B.S., M.S., and Ph.D. degrees in control
science and engineering from Beihang University (Beijing University
of Aeronautics and Astronautics), Beijing, China, in 1984, 1987, and
1991, respectively. He has been a Full Professor with Beihang University
since 1995. He is a Cheung Kong Scholar (Chair Professor), appointed
by the Ministry of Education of China in 1999. His main research interests
include software testing, software reliability, reliable flight control,
and software cybernetics.
\end{IEEEbiography}

\end{document}